\documentclass[times,11pt]{article}

\usepackage{amssymb} 
\usepackage{amsmath} 
\usepackage{latexsym} 
\usepackage{amsthm}
\usepackage{epsf} 
\usepackage[all]{xy} 
\usepackage{epsfig} 
\usepackage{lscape} 
\usepackage{diagrams}  
\usepackage{verbatim}    
\usepackage{times}  
\usepackage{color}

 \newtheorem{theorem}{Theorem}[section]

\theoremstyle{definition}
\theoremstyle{definition}
\theoremstyle{definition}\newtheorem{definition}[theorem]{Definition}
\theoremstyle{definition}
\theoremstyle{definition}
\theoremstyle{definition}
\theoremstyle{definition}
\theoremstyle{definition}
\theoremstyle{definition}
\theoremstyle{definition}

\newcommand{\rem}[1]{\relax} 
\newcommand{\cut}[1]{\relax} 

\renewcommand{\phi}{\varphi}

\newcommand{\bit}{\begin{itemize}} 
\newcommand{\eit}{\end{itemize}\par\noindent} 
\newcommand{\ben}{\begin{enumerate}} 
\newcommand{\een}{\end{enumerate}\par\noindent} 
\newcommand{\beq}{\begin{equation}} 
\newcommand{\eeq}{\end{equation}\par\noindent} 
\newcommand{\beqa}{\begin{eqnarray*}} 
\newcommand{\eeqa}{\end{eqnarray*}\par\noindent}  
\newcommand{\beqn}{\begin{eqnarray}}   
\newcommand{\eeqn}{\end{eqnarray}\par\noindent}

\def\qed{\ifmmode 
          $\Box$ 
         \else 
         {\unskip 
          \nobreak 
          \hfil 
          \penalty50 
          \hskip1em 
          \null 
          \nobreak 
          \hfil 
          $\Box$ 
          \parfillskip=0pt 
          \finalhyphendemerits=0 
          \endgraf} 
         \fi}

\def\II{{\rm I}}

\newcommand{\card}[1]{\,\mid\!\! #1 \!\!\mid}

 \title{Mathematical Foundations for a\\ Compositional  Distributional  Model of Meaning
}
\date{}
\author{Bob Coecke$^\ast$, Mehrnoosh Sadrzadeh$^\ast$, Stephen Clark$^\dag$\\ \\
{\footnotesize coecke, mehrs@comlab.ox.ac.uk -- stephen.clark@cl.cam.ac.uk}\\ \\
$^\ast$Oxford University Computing Laboratory\\
$^\dag$University of Cambridge Computer Laboratory
}
 
\begin{document}

\maketitle 
  
\begin{abstract}

We propose a mathematical framework for a unification of the
distributional theory of meaning in terms of vector space models, and
a compositional theory for grammatical types, for which we rely on the
algebra of Pregroups, introduced by Lambek. This mathematical
\mbox{framework} enables us to compute the meaning of a well-typed sentence
from the meanings of its constituents.  Concretely, the type reductions
of Pregroups are `lifted' to morphisms in a category, a procedure that
transforms meanings of constituents into a meaning of the (well-typed)
whole. Importantly, meanings of whole sentences live in a single
space, independent of the grammatical structure of the sentence.
Hence the inner-product can be used to compare meanings of arbitrary
sentences, as it is for comparing the meanings of words in the
distributional model.  The mathematical structure we employ admits a
purely diagrammatic calculus which exposes how the information flows
between the words in a sentence in order to make up the meaning of the
whole sentence.  A variation of our `categorical model' which involves
constraining the scalars of the vector spaces to the semiring of
Booleans results in a Montague-style Boolean-valued semantics.

\end{abstract}

\section{Introduction}
The symbolic~\cite{Montague} and distributional~\cite{Schutze}
theories of meaning are somewhat orthogonal with competing pros and
cons: the former is compositional but only qualitative, the latter is
non-compositional but quantitative. For a discussion of these two
competing paradigms in Natural Languge Processing see \cite{Gazdar}.
Following \cite{Smolensky} in the context of Cognitive Science, where
a similar problem exists between the connectionist and symbolic models
of mind, \cite{ClarkPulman} argued for the use of the tensor product
of vector spaces and pairing the vectors of meaning with their roles.
In this paper we will also use tensor spaces and pair vectors with
their grammatical types, but in a way which overcomes some of the
shortcomings of \cite{ClarkPulman}. One shortcoming is that, since
inner-products can only be computed between vectors which live in the
same space, sentences can only be compared if they have the same
grammatical structure.  In this paper we provide a procedure to
compute the meaning of any sentence as a vector within a single
space. A second problem is the lack of a method to compute the vectors
representing the grammatical type; the procedure presented here does
not require such vectors.


The use of Pregroups for analysing the structure of natural languages
is a recent development by Lambek \cite{Lambek6} and builds on his
original Lambek (or Syntactic) calculus \cite{Lambek}, where types
are used to analyze the syntax of natural languages in a simple
equational algebraic setting. Pregroups have been used to analyze the
syntax of a range of different languages, from English and French to
Polish and Persian~\cite{SadrPersian}, and many more; for more
references see ~\cite{CasadioLambek,LambekBook}. 


But what is particularly interesting about Pregroups, and motivates
their use here, is that they share a common structure with vector
spaces and tensor products, when passing to a category-theoretic
perspective.  Both the category of vector spaces, linear maps and the
tensor product, as well as pregoups, are examples of so-called compact
closed categories. Concretely, Pregroups are posetal instances of the
categorical logic of vector spaces, where juxtaposition of types
corresponds to the monoidal tensor of the monoidal category.  The
mathematical structure within which we compute the meaning of
sentences will be a compact closed category which combines the two
above.  The meanings of words are vectors in vector spaces, their
grammatical roles are types in a Pregroup, and tensor product of
vector spaces paired with the Pregroup composition is used for the
composition of (meaning, type) pairs.

Type-checking is now an essential fragment of the overall categorical
logic, and the reduction scheme to verify grammatical correctness of
sentences will not only provide a statement on the well-typedness of a
sentence, but will also assign a vector in a vector space to each
sentence.  Hence we obtain a theory with both Pregroup analysis and
vector space models as constituents, but which is inherently
compositional and assigns a meaning to a sentence given the meanings
of its words.  The vectors $\overrightarrow{s}$ representing the
meanings of sentences all live in the same meaning space $S$.  Hence
we can compare the meanings of any two sentences
$\overrightarrow{s},\overrightarrow{t}\in S$ by computing their
inner-product $\langle\overrightarrow{s}|\overrightarrow{t}\rangle$.

Compact closed categories admit a beautiful purely diagrammatic
calculus that simplifies the meaning computations to a great extent.
They also provide reduction diagrams for typing sentences; these allow
for a high level comparison of the grammatical patterns of sentences
in different languages \cite{SadrAAAI}. This diagrammatic structure,
for the case of vector spaces, was recently exploited by Abramsky and
the second author to expose the \em flows of information \em withing
quantum information protocols \cite{AbrCoe2,Kindergarten,Coe}.  Here,
they will expose the flow of information between the words that make
up a sentence, in order to produce the meaning of the whole sentence.
Note that the connection between linguistics and physics was also
identified by Lambek himself \cite{LambekLNP}.

Interestingly, a Montague-style Boolean-valued semantics emerges as a
simplified variant of our setting, by restricting the vectors to range
over $\mathbb{B} = \{0,1\}$, where sentences are simply true or
false. Theoretically, this is nothing but the passage from the
category of vector spaces to the category of relations as described in
\cite{Cats}.  In the same spirit, one can look at vectors ranging
over $\mathbb{N}$ or $\mathbb{Q}$ and obtain degrees or probabilities
of meaning.  As a final remark, in this paper we only set up our
general mathematical framework and leave a practical implementation
for future work.

\section{Two `camps' within computational linguistics}

We briefly present the two domains of Computational Linguistics which
provide the linguistic background for this paper, and refer the reader
to the literature for more details.

\subsection{Vector space models of meaning} 
The key idea behind vector space models of
meaning~\cite{Schutze} can be summed up by Firth's oft-quoted dictum
that ``you shall know a word by the company it keeps''. The basic idea is that the meaning of a word can be determined by the words which
appear in its contexts, where context can be a simple $n$-word window,
or the argument slots of grammatical relations, such as the direct
object of the verb {\em eat}.  Intuitively, {\em cat} and {\em dog} have
similar meanings (in some sense) because cats and dogs sleep, run,
walk; cats and dogs can be bought, cleaned, stroked; cats and dogs can
be small, big, furry. This intuition is reflected in text because {\em
cat} and {\em dog} appear as the subject of {\em sleep}, {\em run},
{\em walk}\/; as the direct object of {\em bought}, {\em cleaned},
{\em stroked}\/; and as the modifiee of {\em small}, {\em big}, {\em
furry}. 

Meanings of words can be represented as vectors in a high-dimensional
``meaning space'', in which the orthogonal basis vectors are
represented by context words. To give a simple example, if the basis
vectors correspond to {\em eat}, {\em sleep}, and {\em run}, and the
word {\em dog} has {\em eat} in its context 6 times (in some text),
{\em sleep} 5 times, and {\em run} 7 times, then the vector for {\em
  dog} in this space is (6,5,7).\footnote{In practice the counts are
  typically weighted in some way to reflect how informative the
  contextual element is with respect to the meaning of the target
  word.}  The advantage of representing meanings in this way is that
the vector space gives us a notion of distance between words, so that
the inner product (or some other measure) can be used to determine how
close in meaning one word is to another.  Computational models along
these lines have been built using large vector spaces (tens of
thousands of context words/basis vectors) and large bodies of text (up
to a billion words in some experiments).  Experiments in constructing
thesauri using these methods have been relatively successful. For
example, the top 10 most similar nouns to {\em introduction},
according to the system of \cite{Curran}, are {\em launch,
  implementation, advent, addition, adoption, arrival, absence,
  inclusion, creation}.

The other main approach to representing lexical semantics is through
an ontology or semantic network, typically manually created by
lexicographers or domain experts. The advantages of vector-based
representations over hand-built ontologies are that:

\begin{itemize}

\item they are created objectively and
automatically from text; 

\item they allow the representation of gradations
of meaning; 

\item they relate well to experimental evidence indicating that
the human cognitive system is sensitive to distributional information
\cite{Saffran:ea:99,Spence:Owens:90}.

\end{itemize}

Vector-based models of word meaning have been fruitfully applied to
many language processing tasks. Examples include lexicon acquisition
\cite{grefenstette94:explorations,Lin:98}, word sense discrimination
and disambiguation \cite{Schutze,McCarthy:ea:04}, text
segmentation \cite{choi01:latent}, language modelling
\cite{Bellegarda:00}, and notably document retrieval
\cite{Salton:ea:75}. Within cognitive science, vector-based models
have been successful in simulating a wide variety of semantic
processing tasks ranging from semantic priming
\cite{lundburgess1996,landauerdumais1997,mcdonald2000} to episodic
memory \cite{Griffiths:Steyvers:07}, and text comprehension
\cite{landauerdumais1997,Foltz:ea:98,Lee:ea:05}. Moreover, the cosine
similarities obtained within vector-based models have been shown to
substantially correlate with human similarity judgements
\cite{mcdonald2000} and word association norms
\cite{denhierelemaire2004,Griffiths:Steyvers:07}.

\subsection{Algebra of Pregroups as a type-categorial logic} 

We provide a brief overview of the algebra of Pregroups from the existing literature and refer the reader for more details to \cite{Lambek6,Lambek7, LambekBook,Busz}. 

A \em partially ordered monoid \em $(P, \leq, \cdot, 1)$ is a partially ordered set, equipped with a monoid multiplication $-\cdot-$ with unit $1$,  where for $p,q,r\in P$, if  $p\leq q$ then we have $r\cdot p\leq r\cdot q$ and  $p\cdot r\leq q\cdot r$. 
A \em Pregroup \em $(P, \leq, \cdot, 1, (-)^l, (-)^r)$ is a partially ordered monoid whose each element $p \in P$ has a \em left adjoint \em $p^l$ and a \em right adjoint \em  $p^r$, i.e. the following hold:
\[
p^l \cdot p \leq 1 \leq p \cdot p^l\qquad \mbox{\rm and} \qquad p \cdot p^r \leq 1 \leq p^r \cdot p\,.
\]
Some properties of interest in a Pregroup are:
\begin{itemize}
\item Adjoints are unique.
\item Adjoints are order reversing: $p \leq q \implies q^r \leq p^r \, \text{and} \, q^l \leq p^l$.
\item The unit is  self adjoint: $1^l = 1 = 1^r$.
\item Multiplication is self adjoint:$(p \cdot q)^r = q^r \cdot p^r$ and $ (p \cdot q)^l = q^l \cdot p^l$.
\item Opposite adjoints annihilate each other: $(p^l)^r = p = (p^r)^l$.
\item Same adjoints iterate: $p^{ll} p^l \leq \!1\! \leq p^r p^{rr}, p^{lll} p^{ll} \leq \!1\! \leq p^{rr} p^{rrr},\ldots\,.$
\end{itemize}
An example of a Pregorup from arithmetic is  the set of all monotone unbounded 
maps on integers $f \colon \mathbb{Z} \to \mathbb{Z}$. In this Pregroup,   function composition is the monoid multiplication and the identity map is its unit, the underlying order on integers lifts to an order on the maps whose Galois adjoints are their Pregroup adjoints, defined:
\[
f^l(x) = min \{y \in \mathbb{Z} \mid x \leq f(y)\} \qquad f^r(x) = max \{y\in \mathbb{Z} \mid f(y) \leq x\}
\]
Recall that a Lambek Calculus $(P, \leq, \cdot, 1, / , \setminus)$ is also a partially  ordered monoid, but there it is  the monoid multiplication that has a right $-\setminus -$ and a left $-/-$ adjoint. Roughly speaking,  the passage from Lambek Calculus to Pregroups can be thought of  as replacing the two adjoints of the monoid 
multiplication with the two adjoints of the elements.  
One can define a translation between a Lambek Calculus and a Pregroup by sending $(p\!\setminus \!q)$ to $(p^r \cdot q)$ and $(p/q)$ to $(p \cdot q^l)$, and via the lambda calculus correspondence of the former think of the adjoint types of a Pregroup as function arguments.

Pregroups formalize grammar of natural languages in the same way as  type-categorial logics do. One starts by fixing a set of  basic grammatical roles and a partial ordering between them,  then freely
generating a Pregroup of these types, the existence of which have been proved. In this paper, we present  two examples from English: positive and negative transitive sentences\footnote{By a negative sentence we mean one with a negation operator, such as {\em not}, and a positive sentence one without a negation operator.}, for which we fix the following  basic types:

\smallskip
\begin{tabular}{lcl}
$n$: noun & $\quad$ &$s$: declarative statement\\
$j$: infinitive of the verb &$\quad$ &$\sigma$:  glueing type
\end{tabular}

\smallskip\noindent
Compound types are formed from these by taking adjoints and juxtaposition.  A type (basic or compound)  is assigned to each word of the dictionary.  We define that  if the juxtaposition of the types of the words within a  sentence reduces to the basic type $s$,  then the sentence is grammatical.  It has been shown 
 that this procedure is decidable.  In what follows we use an arrow $\to$ for $\leq$ and drop the $\cdot$
between juxtaposed types. The example sentence ``John likes Mary", has the following type assignment\footnote{The brackets are only for the purpose of clarity of  exposition and are not part of the mathematical presentation.}:
\begin{center}
\begin{tabular}{ccc}
John & likes & Mary\\\
$n$  & $(n^r s n^l)$ & $n$ 
\end{tabular}
\end{center}
and it is grammatical because of the following reduction:
\[
n (n^r s n^l) n  \to   1  s n^l n  \to 1 s 1 \to s
\] 
Reductions are depicted diagrammatically, that of the  above is:

\vspace{2mm}
\begin{center}
\begin{minipage}{7cm} 
\begin{picture}(50,50)(150,150)
\put(210,194){$n$}
\put(235,194){$n^r$}
\put(250,194){$s$}
\put(262,194){$n^l$}
\put(289,194){$n$}
\end{picture}
{\epsfig{figure=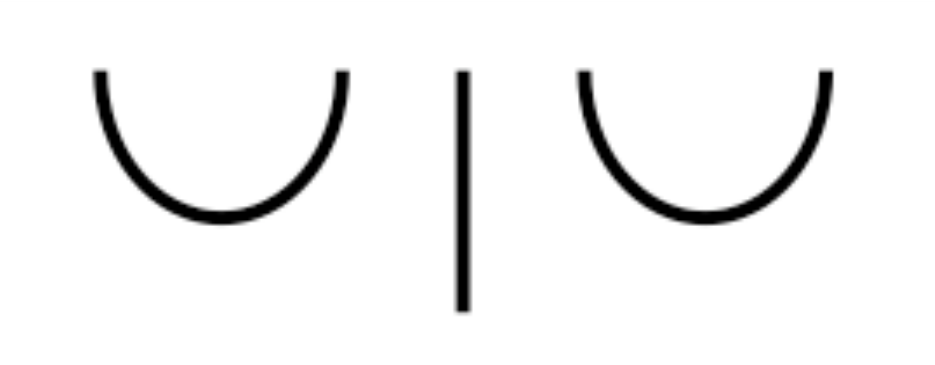,width=100pt}} 
\end{minipage}
\end{center}
Reduction diagrams depict the grammatical structure of sentences in one dimension, as opposed to the two dimensional trees of type-categorial logics.  This feature becomes useful in applications such as comparing the grammatical patterns of different
languages; for some examples see~\cite{SadrAAAI}. 

We type the negation of the above sentence  as follows:
\begin{center}
\begin{tabular}{ccccc}
John & does & not & like & Mary\\
$n$  & $(n^r s j^l \sigma)$ & $(\sigma^r j j^l \sigma)$& $(\sigma^r j n^l)$ & $n$ 
\end{tabular}
\end{center}
which is grammatical because of the following reduction:
\[
n \, (n^r s j^l \sigma)\,  (\sigma^r j j^l \sigma) \, (\sigma^r j n^l)\,  n  \to s
\] 
depicted diagrammatically as follows:
\vspace{2mm}
\begin{center}
\begin{minipage}{7cm}
\begin{picture}(15,15)(150,148)
\put(172,190){$n$}
\put(193,190){$n^r$}
\put(202,190){$s$}
\put(209,190){$j^l$}
\put(216,190){$\sigma$}
\put(236,190){$\sigma^r$}
\put(245,190){$j$}
\put(249,190){$j^l$}
\put(256,190){$\sigma$}
\put(277,190){$\sigma^r$}
\put(286,190){$j$}
\put(291,190){$n^l$}
\put(312,190){$n$}
\end{picture}
{\epsfig{figure=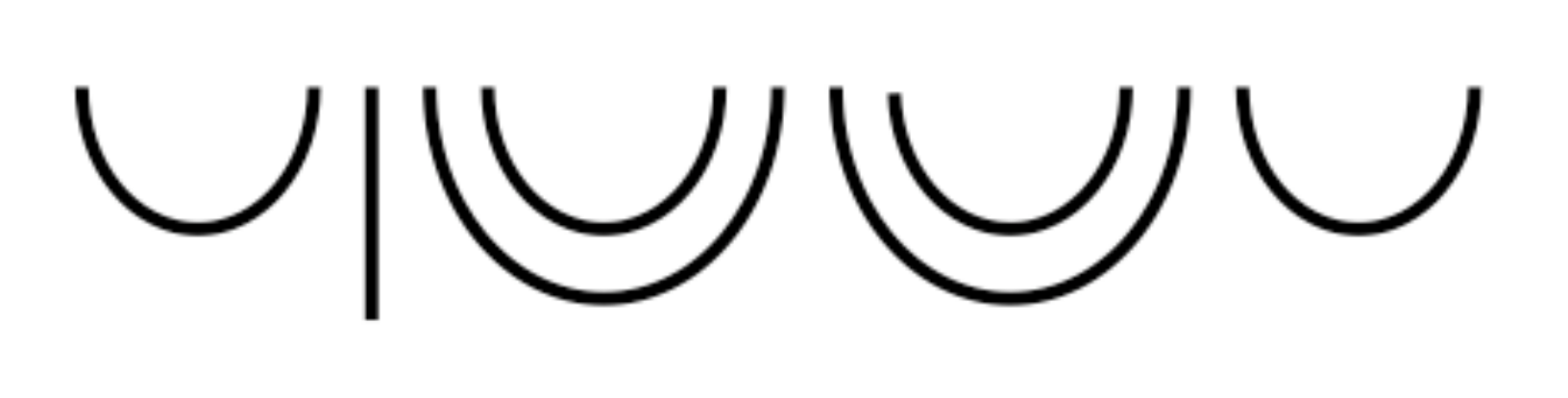,width=157pt}}
\end{minipage}
\end{center}

The types used here for ``does" and ``not" are not the original ones,
e.g. as suggested in \cite{LambekBook}, but are rather obtainable
from the procedure later introduced in \cite{Preller}.  The
difference between the two is in the use of the glueing types; once
these are deleted from the above, the original types are
retrieved. The motivation behind introducing these glueing types is
their crucial role in the development of a discourse semantics for
Pregroups \cite{Preller}. Our motivation, as will be demonstrated in
section 4, is that these allow for the information to flow and be
acted upon in the sentence and as such assist in constructing the
meaning of the whole sentence. Interestingly, we have come to realize
that these new types can also be obtained by translating into the
Pregroup notation the types of the same words from a type-categorial
logic approach, up to the replacement of the intermediate $n$'s with
$\sigma$'s.

\section{Modeling a language in a concrete category}


Our mathematical model of language will be category-theoretic.  Category theory is usually  not conceived as the most evident part of mathematics, so let us briefly state why this passage is essential.
The reader may consult the category theory tutorial \cite{CatsII} which covers the background on the kinds of categories that are relevant here.  Also the survey of graphical languages for monoidal categories \cite{SelingerII}  could be useful -- note that Selinger refers to  \em `non-commutative' compact closed categories \em as (both left and right) \em planar autonomous categories\em. So why do we use categories?
\ben
\item  The passage from $\{\mbox{true, false}\}$-valuations (as in Montague semantics) to quantitative meaning spaces requires a mathematical structure that can store this additional information, but which at the same time retains the compositional structure.  Concrete monoidal categories do exactly that: 
\bit
\item the axiomatic structure, in particular the monoidal tensor,  captures compositionality;
\item the concrete objects and corresponding morphisms enable the encoding of the particular model of meaning one uses, here vector spaces.
\eit
\item The structural morphisms of the particular categories that we consider, compact closed categories, will be the basic building blocks to construct the morphisms that represent the `from-meaning-of-words-to-meaning-of-a-sentence'-process. 
\item Even in a purely syntactic setting, the lifting to categories will allow us to reason about the grammatical structures of different sentences as first class citizens of the formalism. This will enable us to provide more than just a yes-no answer about the grammatical structure of a phrase, i.e. if it is grammatical or not. As such,  the categorical setting will, for instance, allow us to distinguish and reason about  ambiguities in grammatical sentences, where their different grammatical structures gives rise to different meaning interpretations. 
\een
We first briefly recall the basic notions of the theory of monoidal categories, before explaining in more detail what we mean by this `from-meaning-of-words-to-meaning-of-a-sentence'-process. 
 
\subsection{Monoidal categories}\label{sec:monoidalcats}
  
Here we consider the non-symmetric case of a compact closed category, non-degenerate Pregroups being examples of essentially non-commutative compact closed categories.  
The formal definition of monoidal categories is somewhat involved. It
does admit an intuitive operational interpretation and an elegant,
purely diagrammatic calculus.  A (strict) 
monoidal category ${\bf C}$ requires the following data and axioms:
\bit
\item a family $|{\bf C}|$ of \em objects\em;
\bit
 \item     for each ordered pair of objects $(A,B)$ a corresponding set  ${\bf C}(A,B)$ of \em morphisms\em; it is convenient to abbreviate \hbox{$f\in{\bf C}(A,B)$} by $f: A\to B$;

 \item      for each ordered triple of objects $(A,B, C)$, each \hbox{$f:A\to B$}, and $g:B\to C$,  there is a \em sequential composite \em \hbox{$g\circ f:A\to C$}; we moreover require that:
\[
(h\circ g)\circ f=h\circ (g\circ f)\,;
\]
 \item for each object $A$ there is an \em identity morphism \em \hbox{$1_A:A\to A$;} for $f: A\to B$ we moreover require that:
\[
f\circ 1_A= f\qquad \mbox{\rm and} \qquad 1_B\circ f=f\,;
\]
\eit
\item for each ordered pair of objects $(A,B)$ a \em composite object
\em \hbox{$A\otimes B$}; we moreover require that:
\beq\label{eq:assoc}
(A\otimes B)\otimes C=A\otimes (B\otimes C)\,;
\eeq
\item there is a \em unit object \em $\II$ which
satisfies:
\beq\label{eq:unit}
\II\otimes A=A=A\otimes \II\,;
\eeq
\item for each ordered pair of morphisms $(f:A\to C, g: B\to D)$ a \em parallel composite \em $f\otimes g:A\otimes B\to C\otimes D$; we moreover require  \em bifunctoriality \em i.e.
\beq\label{eq;bifunct}
(g_1\otimes g_2)\circ(f_1\otimes f_2)=(g_1\circ f_1)\otimes (g_2\circ f_2)\,.
\eeq
\eit
There is a very intuitive operational interpretation of monoidal
categories.  We think of the objects as \em types of systems\em.  We
think of a morphism $f:A\to B$ as a \em process \em which takes a
system of type $A$  as input and provides a system of type $B$ as output, i.e.~given any state $\psi$ of the system of type $A$, it produces a state $f(\psi)$ of the system of type $B$.  Composition of morphisms is sequential application of processes.  The compound type $A\otimes B$ represents \em joint systems\em.  We think of $\II$ as the trivial system, which can be either `nothing' or `unspecified'.
More on this intuitive interpretation can be found in \cite{Cats,CatsII}.

Morphisms $\psi:\II\to A$ are called \em elements \em of
$A$.  At first this might seem to be a double use of terminology: if $A$ were to be a set, then $x\in A$ would be an element, rather than some function $x:\II\to A$. However, one easily sees that elements in $x\in A$ are in bijective correspondence with functions $x:\II\to A$ provided one takes $\II$ to be a singleton set.  The same holds for vectors $\overrightarrow{v}\in V$, where $V$ is a vector space, and linear maps $\overrightarrow{v}:\mathbb{R}\to V$.  In this paper we take the liberty to jump between these two representations of a vector $\overrightarrow{v}\in V$, when using them to represent meanings.

In the standard definition of monoidal categories  the `strict'
equality of eqs. (\ref{eq:assoc},\ref{eq:unit}) is not required but rather the existence of a \em natural isomorphism \em between $(A\otimes B)\otimes C$ and $A\otimes (B\otimes C)$. We assume
strictness in order to avoid  \em coherence conditions\em.  This simplification is justified by the fact
that each monoidal category is categorically equivalent to a strict
one, which is obtained by imposing appropriate congruences.  Moreover, the graphical language which we introduce below represents (free) strict monoidal categories.  This issue is discussed in detail in \cite{CatsII}.

So what is particularly interesting about these monoidal categories is indeed  that they admit a graphical calculus in the following sense \cite{SelingerII}:
\begin{quote}
\em An equational statement between morphisms in a monoidal category is provable from the axioms of monoidal categories if and only if it is derivable in the graphical language\em.
\end{quote}
This fact moreover does not only hold for ordinary monoidal categories, but also for many kinds that have additional structure, including the compact closed categories that we will consider here.

\paragraph{Graphical language for monoidal categories.} In the graphical calculus for monoidal categories we depict morphisms
by boxes, with incoming and outgoing wires labelled by the
corresponding types, with sequential composition depicted by
connecting matching outputs and inputs, and with parallel composition
depicted by locating boxes side by side. For example, the morphisms
\[
1_A  \qquad f  \qquad g\circ f \qquad 1_A\otimes 1_B  \qquad f\otimes 1_C  \qquad f\otimes g  \qquad (f\otimes g)\circ h
\]
are depicted as follows in a top-down fashion: 
\par\vspace{1mm}\noindent
\hspace{-0.1cm}\begin{minipage}[b]{1\linewidth}
{\epsfig{figure=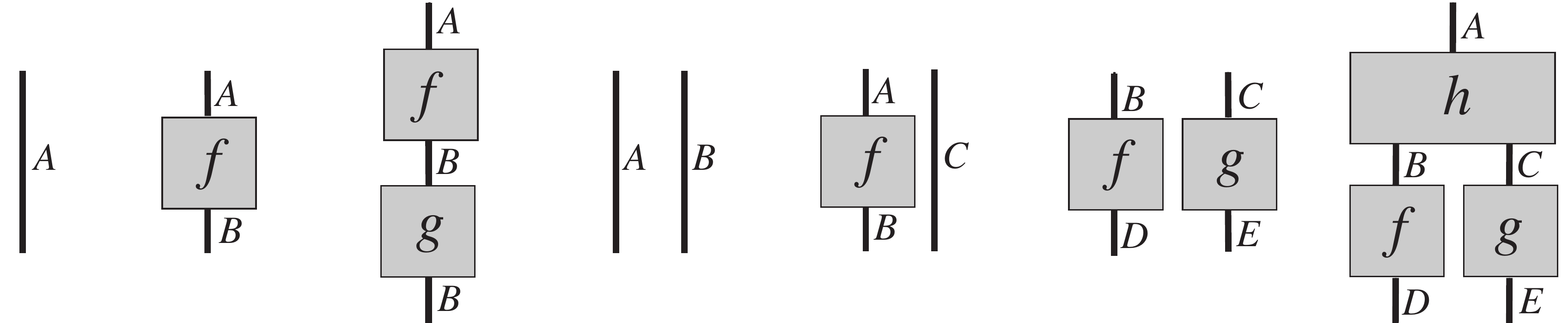,width=290pt}}      
\end{minipage}
\par\vspace{0mm}\noindent
When representing morphisms in this manner by boxes, eq.(\ref{eq;bifunct}) comes for free \cite{CatsII}! 

The unit object $\II$ is represented by `no wire'; for example
\[
\psi:\II\to A\quad\qquad \pi:A\to \II \quad\qquad \pi\circ\psi:\II\to\II
\]
are depicted as:
\par\vspace{0mm}\noindent
\begin{minipage}[b]{1\linewidth}
\centering{\epsfig{figure=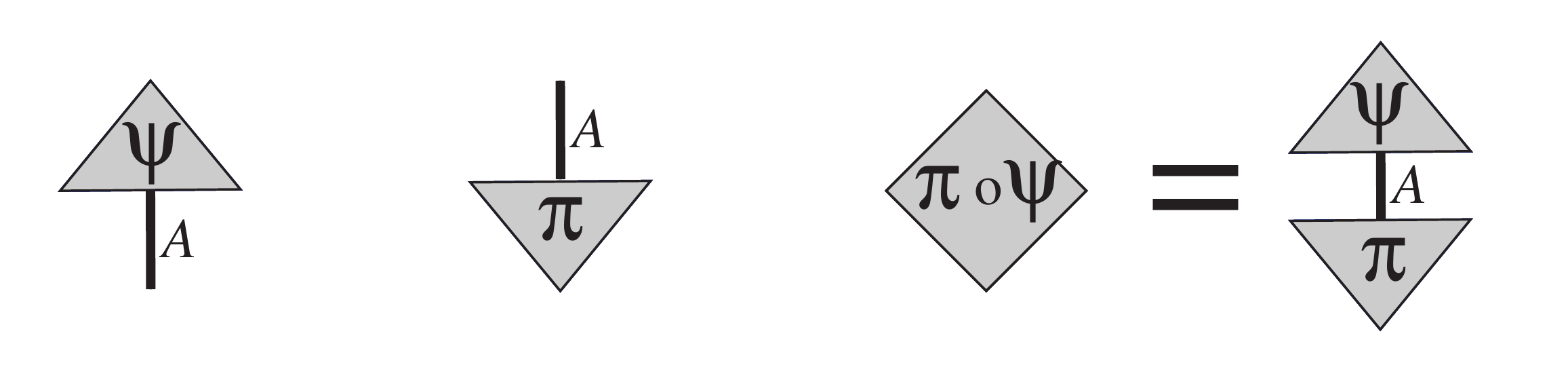,width=230pt}}     
\end{minipage}

\subsection{The `from-meaning-of-words-to-meaning-of-a-sentence' process}

Monoidal categories are widely used to represent processes between
systems of varying types, e.g.~data types in computer programs. The
process which is central to this paper is the one which takes the
meanings of words as its input and produces the meaning of a sentence
as output, within a fixed type $S$ (Sentence) that allows the
representation of meanings of all well-typed sentences.

Diagrammatically we represent it as follows: 
\begin{center}
\epsfig{figure=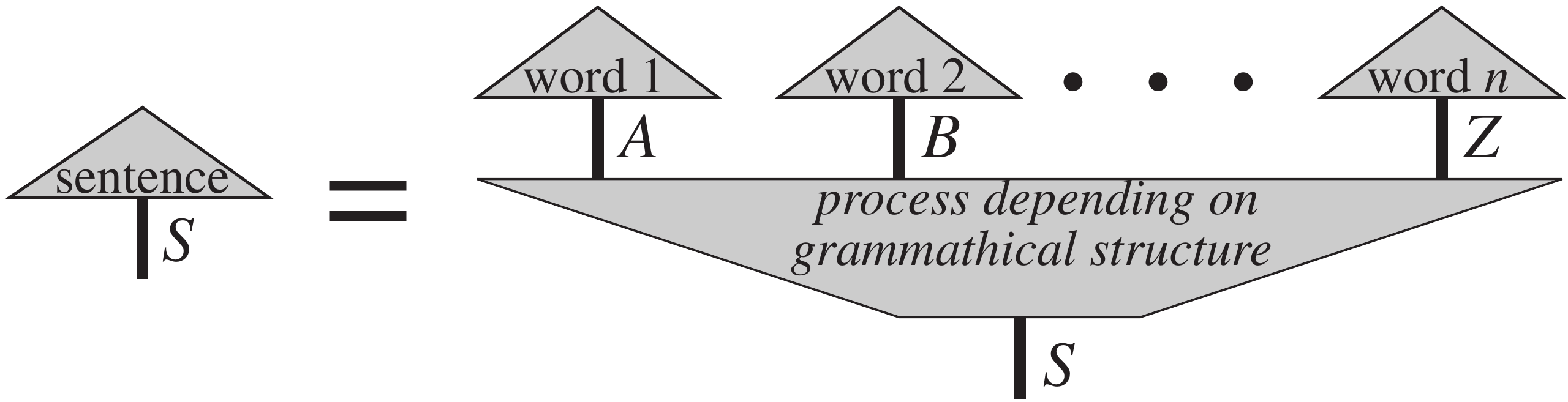,width=230pt}
\end{center} 
where all triangles represent meanings, both of words and sentences. For example, the triangle labeled `word 1' represents the meaning of word 1 which is of grammatical  type $A$, and the triangle labeled `sentence' represents the meaning of the whole sentence. The concatenation (word 1)$\cdot$ \ldots$\cdot$ (word n) is the sentence itself, which is of grammatical type $A\otimes \ldots \otimes Z$, and the way in which the list of meanings of words:
\begin{center}
\epsfig{figure=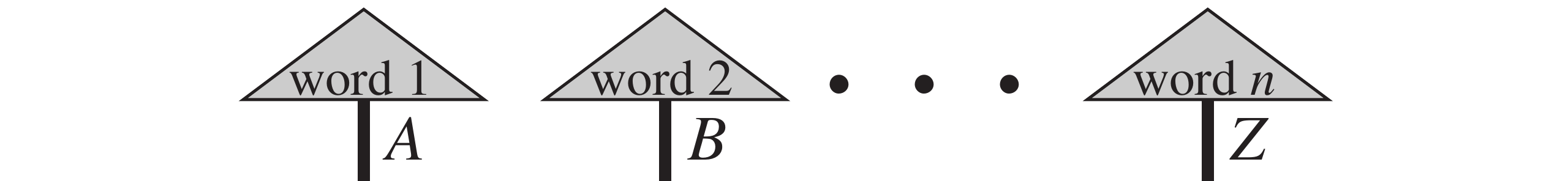,width=230pt}
\end{center} 
 becomes the meaning of a sentence:
\begin{center}
\epsfig{figure=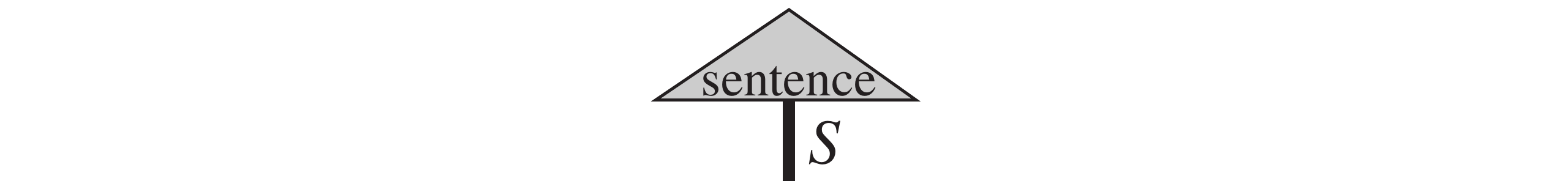,width=230pt}\,
\end{center} 
within the fixed type $S$, is mediated by the grammatical structure. The concrete manner in which grammatical structure performs this role will be explained below.  This method will exploit the common mathematical structure which vector spaces (used to assign meanings to words in a language) and Pregroups (used to assign grammatical structure to sentences) share, namely compact closure.  

\subsection{Compact closed categories}
 
A monoidal category is \em compact closed \em if for each object $A$ there are
also objects $A^r$ and $A^l$, and morphisms
\[
\eta^l:\II\to A\otimes A^l\!
\quad
\epsilon^l:A^l\!\otimes A\to\II
\quad
\,\eta^r\!:\II\to A^r\!\otimes A
\quad
\epsilon^r\!:A\otimes A^r\!\to\II
\]
which satisfy:
\[
(1_A\otimes \epsilon^l)\circ(\eta^l\otimes 1_A)=1_A \ \
\quad
\ \ (\epsilon^r\!\otimes 1_{A})\circ(1_{A}\otimes\eta^r\!)=1_{A}\vspace{1.2mm}
\]
\[
\ (\epsilon^l\otimes 1_{A^l})\circ(1_{A^l}\otimes\eta^l)=1_{A^l}
\quad
(1_{A^r}\otimes \epsilon^r\!)\circ(\eta^r\!\otimes 1_{A^r})=1_{A^r}
\]

Compact closed categories are in a sense orthogonal to cartesian categories, such as the category of sets and functions with the cartesian product as the monoidal structure.  Diagrammatically, in a cartesian category the triangles representing meanings of type $A\otimes B$ could always be decomposed into a triangle   representing meanings of type $A$ and a triangle   representing meanings of type $B$:
\[ 
{{{\bf Cartesian}\over{\bf non\mbox{-}Cartesian}} 
\ =\  
{\epsfig{figure=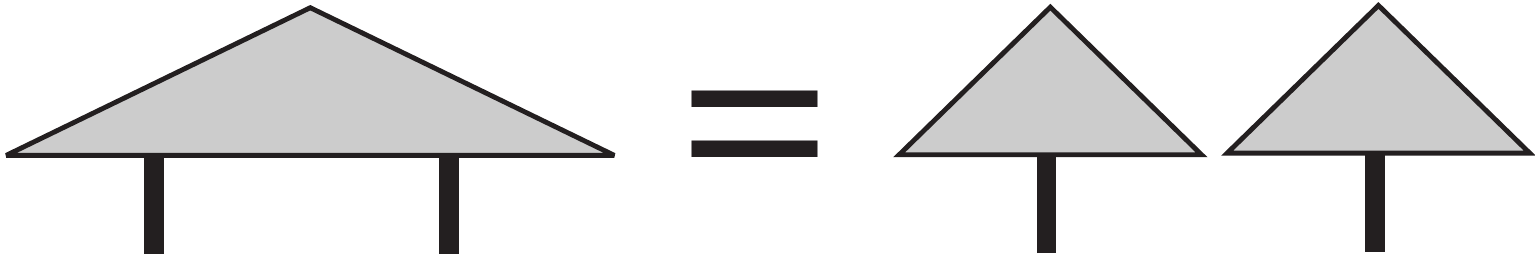,width=160pt}\over\epsfig{figure=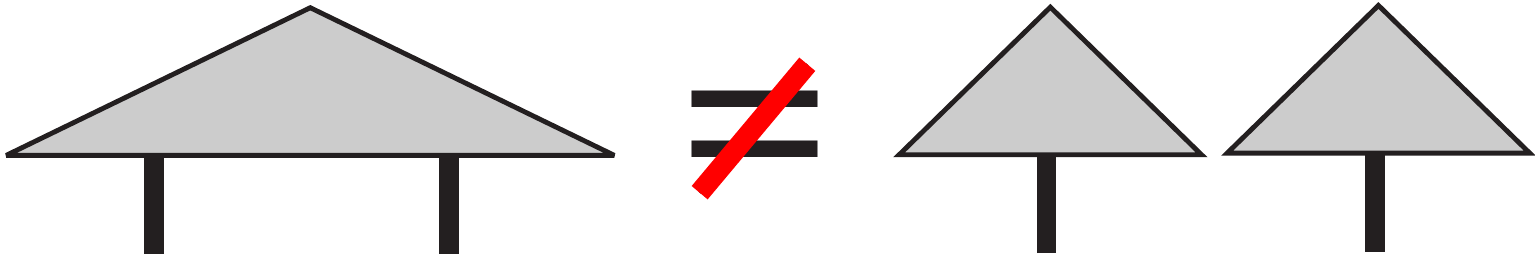,width=140pt}}}
\]
But if we consider a verb, then its grammatical type is $n^r s n^l$, that is, of the form $N\otimes S\otimes N$ within the realm of monoidal categories.  Clearly, to compute the meaning of the whole sentence, the meaning of the verb will need to \em interact \em with the meaning of both the object and subject, so it cannot be decomposed into three disconnected entities:
\begin{center}
\epsfig{figure=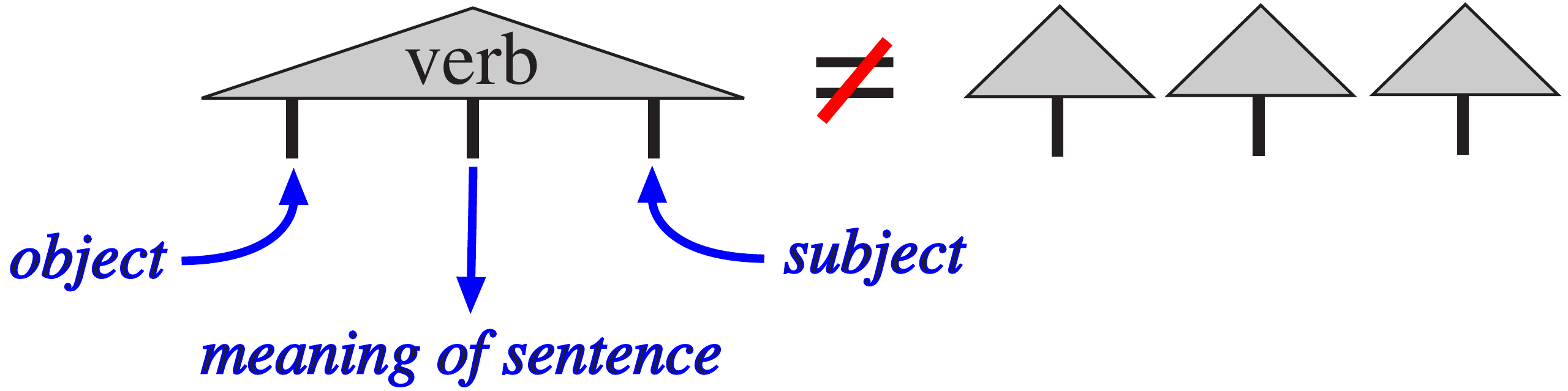,width=250pt}
\end{center} 
In this graphical language, the topology (i.e.~either being connected or not) represents when interaction occurs.  In other words, `connectedness' encodes `correlations'.

That we cannot always decompose  triangles representing meanings of type $A\otimes B$ in compact closed categories can be immediately seen in the graphical calculus of compact closed categories, which explicitly introduces wires between different types, and these will mediate flows of information between words in a sentence.  A fully worked out example of sentences of this type is given in section \ref{sec:verbexample}.

\paragraph{Graphical language for compact closed categories.} 
When depicting the morphisms $\eta^l, \epsilon^l, \eta^r, \epsilon^r$ as (read in a top-down fashion)
\par\vspace{3mm}\noindent
\begin{minipage}[b]{1\linewidth}
\centering{\epsfig{figure=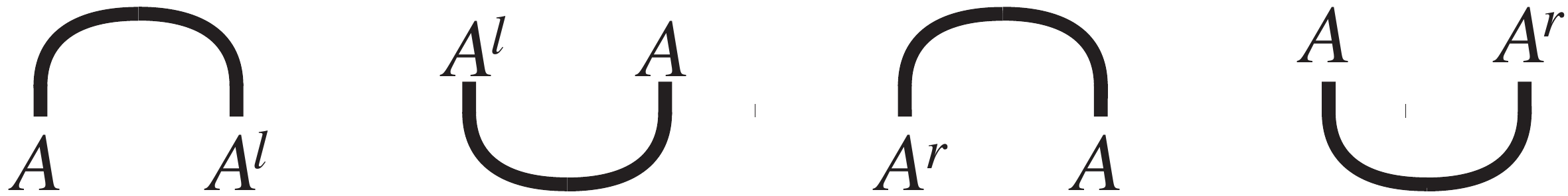,width=220pt}}      
\end{minipage}
\par\vspace{2mm}\noindent
rather than as triangles, the axioms of compact closure simplify to:
\par\vspace{3mm}\noindent
\begin{minipage}[b]{1\linewidth}
\centering{\epsfig{figure=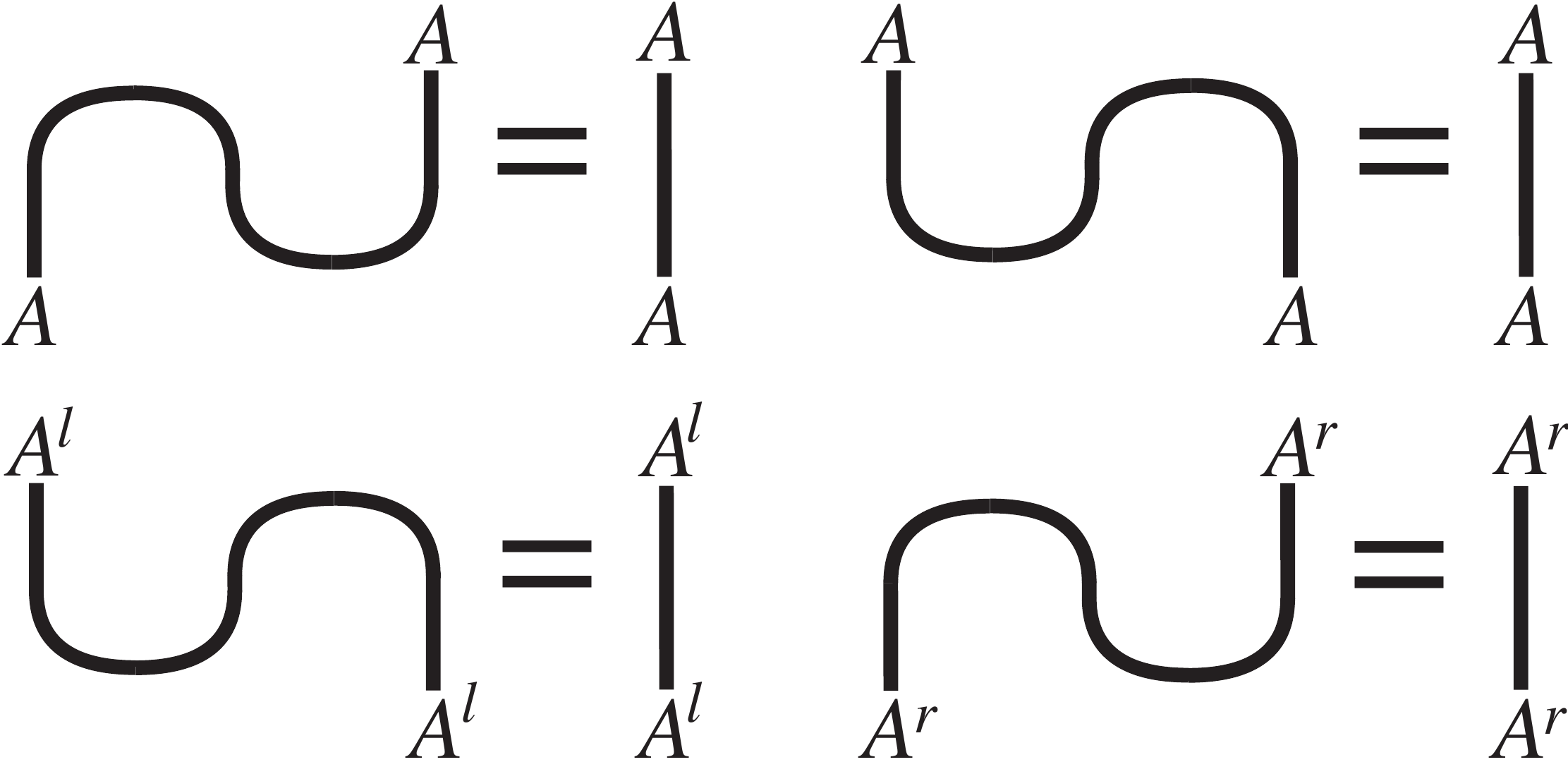,width=200pt}}      
\end{minipage}
\par\vspace{2mm}\noindent
i.e.~they boil down to `yanking wires'.

\paragraph{Vector spaces, linear maps and tensor product as a compact closed category.} 
Let ${\bf FVect}$ be the category which has vector spaces over the base field $\mathbb{R}$ as objects, linear maps as morphisms and the vector space tensor product as the monoidal tensor.  In this category, the tensor is commutative, i.e. $V \otimes W \cong W \otimes V$, and left and right adjoints are the same, i.e. $V^l=V^r$ so we denote either by $V^*$, which is the  identity maps, i.e. $V^* = V$. To simplify the presentation we assume that each vector space comes with an inner product, that is, it is an \em inner-product space\em. For the case of vector space models of meaning this is always the case, since we consider a fixed base, and a fixed base canonically induces an inner-product.  The reader can verify that compact closure arises, given a vector space $V$ with base $\{\overrightarrow{e}_i\}_i$, by setting $V^l=V^r=V$,
\beq \label{def:etavec}
\eta^l=\eta^r\!:\mathbb{R}\to V\otimes V:: 1\mapsto \sum_i \overrightarrow{e}_i\otimes \overrightarrow{e}_i \vspace{-2mm}
\eeq
and
\beq\label{def:epsilonvec}
\epsilon^l=\epsilon^r\!: V\otimes V\!\to\mathbb{R}:: \sum_{ij}c_{ij}\,\overrightarrow{v}_i\otimes\overrightarrow{w}_j\mapsto\sum_{ij}c_{ij}\langle\overrightarrow{v}_i |\overrightarrow{w}_j\rangle\,.
\eeq
In equation~\ref{def:etavec} we have that  $\epsilon^l=\epsilon^r$ is the inner-product extended by linearity to the whole tensor product. 
Recall that if $\{\overrightarrow{e}_i\}_i$ is a base for $V$ and if $\{\overrightarrow{e}'_i\}_i$  is a base for $W$ then $\{\overrightarrow{e}_i\otimes \overrightarrow{e}'_j\}_{ij}$ is a base for $V\otimes W$.  In the base $\{\overrightarrow{e}_i\otimes \overrightarrow{e}_j\}_{ij}$ for $V\otimes V$ the linear map $\epsilon^l=\epsilon^r: V\otimes V\to\mathbb{R}$ has as its matrix the row vector  which has entry $1$ for the base vectors $\overrightarrow{e}_i\otimes \overrightarrow{e}_i$ and which has entry $0$ for the base vectors $\overrightarrow{e}_i\otimes \overrightarrow{e}_j$ with $i\not= j$. The matrix of $\eta^l=\eta^r$ is the column vector obained by transposition. 

In 
eq. (\ref{def:epsilonvec}), the weighted sum $\sum_{ij}c_{ij}\,\overrightarrow{v}_i\otimes\overrightarrow{w}_j$ denotes a typical vector in a tensor space $V \otimes W$,  where $c_{ij}$'s enumerate  all possible  weights for the tensored pair of base vectors $\overrightarrow{v}_i\otimes\overrightarrow{w}_j$.  If in the definition of $\epsilon^l=\epsilon^r$ we apply the restriction that
$\overrightarrow{v}_i=\overrightarrow{w}_i=\overrightarrow{e}_i$,
which we can do if we stipulate that $\epsilon^l=\epsilon^r$ is a
linear map, then it simplifies to
\[
\epsilon^l=\epsilon^r\!: V\otimes V\!\to\mathbb{R}:: \sum_{ij}c_{ij}\,\overrightarrow{e}_i\otimes\overrightarrow{e}_j\mapsto\sum_{i}c_{ii}\,.
\]


\paragraph{A Pregroup as a compact closed category.} 
A Pregroup is an example of a \em posetal category\em, that is, a
category which is also a poset.  For a category this means that for
any two objects there is either one or no morphism between them. In
the case that this morphism is of type $A\to B$ then we write $A\leq
B$, and in the case it is of type $B\to A$ we write $B\leq A$.  The
reader can then verify that the axioms of a category guarantee that
the relation $\leq$ on $|C|$ is indeed a partial order. Conversely,
any partially ordered set $(P,\leq)$ is a category. For `objects'
$p,q,r \in P$ we take $[p \leq q]$ to be the singleton $\{p\leq q\}$ whenever
$p\leq q$, and empty otherwise.  If $p\leq q$ and $q\leq r$ we define
$p\leq r$ to be the composite of the `morphisms' $p\leq q$ and $q\leq
r$.

A partially ordered monoid is a monoidal category with the monoid
multiplication as tensor on objects; whenever $p\leq r$ and $q\leq z$
then we have $p\cdot q\leq r\cdot z$ by monotonicity of monoid multiplication,
and we define this to be the tensor of `morphisms' $[p \leq r]$ and
$[q \leq z]$.  Bifunctoriality, as well as any equational statement
between morphisms in posetal categories, is trivially satisfied, since
there can only be one morphism between any two objects.

Finally, each Pregroup is a compact closed category for
\[
\eta^l=[1 \leq p \cdot p^l] \qquad \epsilon^l=[p^l \cdot p \leq 1]   
\]
\[
\eta^r\!=[1 \leq p^r\! \cdot p] \qquad \epsilon^r\!=[p \cdot p^r\! \leq 1] 
\]
and so the required equations are again trivially satisfied.
Diagrammatically, the under-links representing the type reductions in a Pregroup grammar
are exactly the `cups'
of the compact closed structure.  The symbolic counterpart of the diagram of the reduction of a sentence with a transitive verb
\begin{center}
\begin{minipage}{6cm} 
\begin{picture}(50,50)(150,150)
\put(210,194){$n$}
\put(235,194){$n^r$}
\put(250,194){$s$}
\put(262,194){$n^l$}
\put(289,194){$n$}
\end{picture}
{\epsfig{figure=likesend.pdf,width=100pt}} 
\end{minipage}
\end{center}
is the following  morphism:
\[
\epsilon^r_n\otimes 1_s\otimes \epsilon^l_n:n\otimes n^r\otimes s\otimes n^l\otimes n\to s\,.
\]

\subsection{Categories representing both grammar and meaning}

We have described two aspects of natural language which admit mathematical presentations:
\ben
\item 
 vector spaces can be used to assign meanings to words in a language;
 \item Pregroups can be used to assign grammatical structure to sentences.  
 \een
 When we organize these vector spaces as a monoidal category by also considering linear maps, and tensor products both of vector spaces and linear maps, then  these  two mathematical objects share common structure, namely compact closure.  We can think of these two compact closed structures as two structures that we can \em project \em out of a language, where $P$ is the free  Pregroup generated from the basic types of a natural language:
 \begin{diagram}
&& language &&\\
& \ldTo^{meaning} &  & \rdTo^{grammar}& \\
{\bf FVect} & & & & P
\end{diagram}
 
We aim for a mathematical structure that unifies both of these aspects of language, that is, in which the
compositional structure of Pregroups would lift to the level of assigning meaning to sentences and their constituents, or dually, where the structure of assigning meaning to words comes with a mechanism that enables us to compute the meaning of a sentence.  The compact closed structure of ${\bf FVect}$ alone is too degenerate for this purpose since $A^l=A^r=A$. Moreover, there are canonical isomorphisms $V\otimes W\to W\otimes V$ which translate to posetal categories as $a\cdot b= b\cdot a$, and in general we should not be able to exchange words in a sentence without altering its meaning.  Therefore we have to refine types to retain the full grammatical content obtained from the Pregroup analysis. There is an easy way of doing this: rather than objects in ${\bf FVect}$ we will consider objects in the product category ${\bf FVect}\times P$:
\begin{diagram}
&& language &&\\
& \ldTo^{meaning} & \dTo & \rdTo^{grammar}& \\
{\bf FVect} & \lTo_{\pi_m} & {\bf FVect} \times P & \rTo_{\pi_g\!\!\!\!\!\!} & P
\end{diagram}
Explicitly, ${\bf FVect} \times P$ is the
category which has pairs $(V,a)$ with $V$ a vector space and $a\in P$ a grammatical type 
as objects, and the following pairs as morphisms:
\[
(f:V\to W\,,\,p\leq q)\,,
\]
which we can also write as
\[
(f,\leq): (V,p)\to(W,q).
\]

Note that if $p\not\leq q$ then there are no morphisms of type $(V, p)\to(W,q)$. It is easy to verify that the compact closed structure of ${\bf FVect}$ and $P$ lifts component-wise to one on ${\bf FVect}\times P$.  The structural morphisms in this new category are now:
\[
(\eta^l, \leq): (\mathbb{R} , 1)\to (V\otimes V ,\, p \cdot p^l) \qquad
(\eta^r, \leq): (\mathbb{R} , 1)\to (V\otimes V ,\, p^r \cdot p)
\]
\[
(\epsilon^l, \leq): (V\otimes V ,\, p^l \cdot p)\to (\mathbb{R} , 1)\qquad
(\epsilon^r, \leq): (V\otimes V ,\, p \cdot p^r)\to (\mathbb{R} , 1)
\] 

\subsection{Meaning of a sentence as a morphism in ${\bf FVect} \times P$.}

\begin{definition}\em 
We refer to an object $(W,p)$ of ${\bf Fvect} \times P$ as a  \em meaning space \em. This consists of a vector space $W$ in which the meaning of a word  lives $\overrightarrow{w} \in W$ and the grammatical type $p$ of the word. 
\end{definition}

\begin{definition}\label{meaningdef}\em
We define the  vector $\overrightarrow{w_1 \cdots w_n}$ of  the meaning of a string of words $w_1 \cdots w_n$ to be 
\[
\overrightarrow{w_1 \cdots w_n} := f(\overrightarrow{w_1} \otimes \cdots \otimes \overrightarrow{w_n})
\]
where for   $(W_i, p_i)$  meaning space of  the word $w_i$,  the linear map $f$ 
is built by substituting each $p_i$ in  \hbox{$[p_1 \cdots p_n \leq x]$} with  $W_i$. \end{definition}
Thus for $\alpha = [p_1 \cdots p_n \to x]$ a morphism in ${\bf P}$ and $f = \alpha[p_i\setminus W_i]$ a linear map in ${\bf Fvect}$, the following  is a morphism in  ${\bf Fvect} \times P$:
\[
(W_1 \otimes \cdots \otimes W_n, \,p_1 \cdots p_n) \rTo^{(f, \leq)} (X,x)
\]
We  call $f$ the `from-meaning-of-words-to-meaning-of-a-sentence' map.

\smallskip
According to this formal definition, the  procedure of assigning meaning to a string of words can be roughly described as follows:
\begin{enumerate}
\item  Assign a grammatical type $p_i$ to each word $w_i$ of the string, apply the axioms and rules of the Pregroup grammar to  reduce these types to a simpler type $p_1 \cdots p_n \to x$. If the string of words is a sentence, then the reduced type $x$ should be the basic grammatical type  $s$ of a sentence\footnote{By Lambek's switching lemma~\cite{Lambek6} the  epsilon maps suffice for the grammatical reductions and thus  $x$ already exists in the type of one of the words in the string.}.

\item Assign a vector space to each word of the sentence based on its syntactic type assignment. For the purpose of this paper, we prefer to be flexible with the manner in which these vector spaces are built, e.g. the vector spaces of  the words with basic types like noun may be atomic and  built  according to the usual rules of the distributional model; the vector spaces of the words with compound types  like verbs are tensor spaces.

\item Consider the vector of the meaning of each word in the spaces built above, take their tensor, and apply to it the diagram of the syntactic reduction of the string, according to the meaning spaces of each word. This will provide us with the meaning of the string.  
\end{enumerate}

\subsection{Comparison with the connectionist proposal}
Following the solution of connectionists \cite{Smolensky},  Pulman and the third author argued for the use of tensor products in developing a compositional distributional model of meaning \cite{ClarkPulman}. They suggested that to implement this idea in
linguistics one can, for example, traverse the parse tree of a
sentence and tensor the vectors of the meanings of words with the
vectors of their roles:
\[
(\overrightarrow{\mbox{\em John}}\otimes \overrightarrow{\mbox{\em subj}})\otimes \overrightarrow{\mbox{\em likes}}\otimes(\overrightarrow{\mbox{\em Mary}}\otimes \overrightarrow{\mbox{\em obj}})\,
\]
This vector in the tensor product space should then be regarded as the meaning of the sentence ``John likes Mary."

The tensors  $(\overrightarrow{\mbox{\em John}}\otimes \overrightarrow{\mbox{\em subj}})$ and $(\overrightarrow{\mbox{\em Mary}}\otimes \overrightarrow{\mbox{\em obj}})$ in the above are pure tensors, and thus can be considered as a pair of vectors, i.e. $(\overrightarrow{\mbox{\em John}},  \overrightarrow{\mbox{\em subj}})$ and $(\overrightarrow{\mbox{\em Mary}}, \overrightarrow{\mbox{\em obj}})$. These are pairs of a meaning of a word and its grammatical role, and almost the same as the pairs considered in our approach, i.e. that of a meaning space of each word.  A minor difference is that, in the above,  the grammatical role $\overrightarrow{p}$ is a genuine vector, whereas in our approach this remains a grammatical type.  If needed, our approach can easily be adapted to also allow types to be represented in a vector space.

A more conceptual difference between the two approaches lies in the fact that the above does not assign a grammatical type to the verb, i.e. treats $\overrightarrow{\mbox{\em likes}}$ as a single vector. Whereas in our approach,   the vector of the verb itself lives in a tensor space.

\section{Computing the meaning of example sentences}
In what follows we use the steps above to assign meaning to positive and negative transitive sentences\footnote{For the negative example,  we use the idea and treatment of previous work \cite{PrellerSadr},  in that we use eta maps to interpret the logical meaning of ``does" and ``not", but extend the details of calculations, diagrammatic representations, and corresponding comparisons.}. 

\subsection{Positive Transitive Sentence}\label{sec:verbexample}

A positive sentence with a transitive verb has the Pregroup type $n (n^r\! s n^l) n$. We assume that the meaning spaces of  the subject and object of the sentence are atomic and are given as $(V,n)$ and $(W,n)$. The meaning space of the verb is compound and is given as $(V\otimes~S~\otimes~W,n^r\! s n^l)$. The `from-meaning-of-words-to-meaning-of-a-sentence' linear map $f$  is the linear map which realizes the following structural morphism in ${\bf FVect} \times P$:
\[
\big (V \otimes T \otimes W\,, n (n^r\! s n^l) n \big) \rTo^{(f,\leq)} (S, s)\,,
\]
and arises from a syntactic reduction map; in this case we obtain: 
\[
f= \epsilon_V \otimes 1_S \otimes \epsilon_W:    V \otimes (V \otimes S \otimes W) \otimes W\to S\,.
\]
Noting the isomorphism $V \otimes S \otimes W \cong V \otimes W \otimes S \cong V^* \otimes W^* \to S$ obtained from the commutativity of tensor in the   ${\bf FVect}$   and that $V^* = V$ and $W^* = W$ therein, and the universal property of the tensor with respect to product, we can think about the meaning space of a verb $V \otimes W \otimes S$ as a function space $V \times W \to S$.  So the meaning vector of each transitive verb can be thought of as a function that inputs a subject from $V$ and an object from $W$ and outputs a sentence in $S$.  

In the graphical calculus, the linear map of meaning is depicted as follows: 
\begin{center}
\epsfig{figure=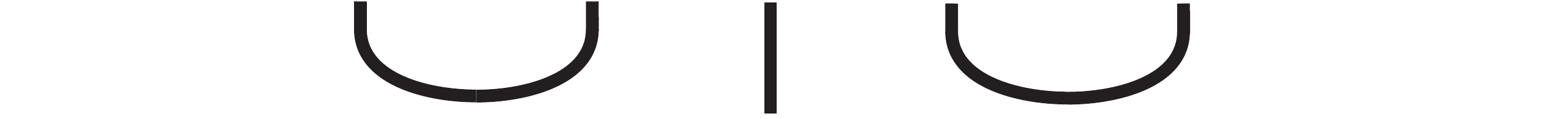,width=250pt}
\end{center} 
The matrix of $f$ has $dim(V)^2\times dim(S)\times dim(W)^2$ columns
and $dim(S)$ rows, and its entries are either $0$ or $1$.  When applied to the vectors of the meanings of the words, i.e.  $f(\overrightarrow{v}\otimes \overrightarrow{\Psi} \otimes
\overrightarrow{w})\in S$ for $\overrightarrow{v}\otimes
\overrightarrow{\Psi} \otimes \overrightarrow{w}\in V \otimes S
\otimes W$ we obtain, diagrammatically:
\begin{center}
\epsfig{figure=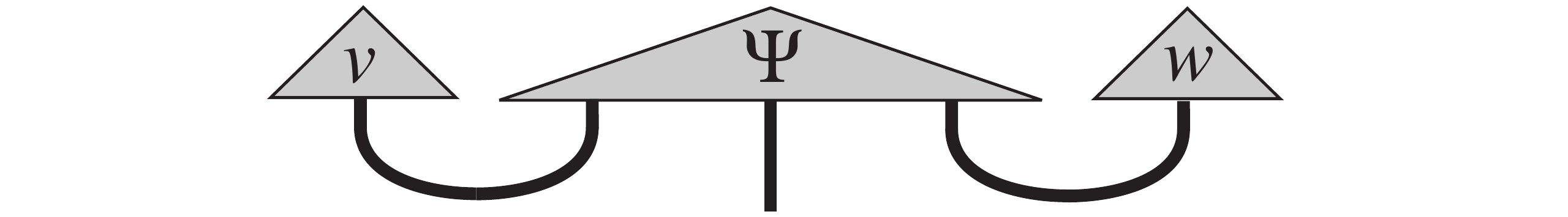,width=250pt}
\end{center} 
This map can be expressed in terms of the inner-product as follows. Consider the typical vector in the tensor space which represents the type of verb:
\[
\Psi=
\sum_{ijk}c_{ijk}\overrightarrow{v}_i\otimes\overrightarrow{s}_{\!j}\otimes\overrightarrow{w}_k
\in V \otimes S \otimes W
\]
then 
\beqa
f(\overrightarrow{v}\otimes \overrightarrow{\Psi} \otimes \overrightarrow{w}) &=& \epsilon_V \otimes 1_S \otimes \epsilon_W(\overrightarrow{v}\otimes \overrightarrow{\Psi} \otimes \overrightarrow{w})\\
\!\!&=&\!\!
\sum_{ijk}c_{ijk}\langle\overrightarrow{v} | \overrightarrow{v}_i\rangle \overrightarrow{s}_{\!j} 
\langle\overrightarrow{w}_k |\overrightarrow{w} \rangle \\ 
\!\!&=&\!\!
\sum_j\left(\sum_{ik}c_{ijk}\langle\overrightarrow{v} | \overrightarrow{v}_i\rangle 
\langle\overrightarrow{w}_k |\overrightarrow{w} \rangle\right)\overrightarrow{s}_{\!j}\,.
\eeqa
This vector is the meaning of the sentence of type $n (n^r\! s
n^l) n$, and assumes as given the meanings of its constituents $\overrightarrow{v}\in V$,
$\overrightarrow{\Psi}\in T$ and $\overrightarrow{w}\in W$, obtained from data using some suitable method. 

Note that, in Dirac notation, $f(\overrightarrow{v}\otimes \overrightarrow{\Psi} \otimes
\overrightarrow{w})$ is written as:
\[
\bigl(\langle \epsilon^r_V|\otimes 1_S\otimes\langle \epsilon^r_V|\bigr)\bigm|\!\!\overrightarrow{v}\otimes \overrightarrow{\Psi} \otimes \overrightarrow{w}\bigr\rangle.
\]
Also, the diagrammatic calculus tells us that:
\begin{center}
\hspace{-0.2cm}\epsfig{figure=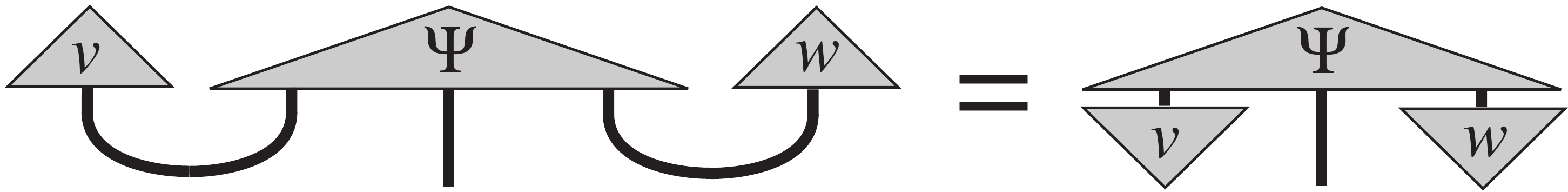,width=280pt}
\end{center} 
where the reversed triangles are now the corresponding Dirac-bra's, or in vector space terms, the corresponding functionals in the dual space.  This simplifies the expression that we need to compute to:
\[
(\langle \overrightarrow{v}|\otimes 1_S\otimes \langle \overrightarrow{v}|)|\overrightarrow{\Psi}\rangle
\]

As mentioned in the introduction, our focus in this paper is not on
how to practically exploit the mathematical framework, which would require substantial further research,
but to expose the mechanisms
which govern it.  To show that this particular computation (i.e.~the `from-meaning-of-words-to-meaning-of-a-sentence'-process)  does indeed produce a vector which captures the meaning of a sentence,
we explicitly compute $f(\overrightarrow{v}\otimes
\overrightarrow{\Psi} \otimes \overrightarrow{w})$ for some
simple examples, with the intention of providing the reader
with some insight into the underlying mechanisms and how the approach
relates to existing frameworks.  

\paragraph{Example 1.}{\bf One Dimensional Truth-Theoretic Meaning.}  Consider the sentence 
\beq\label{sentence}
\mbox{\rm\it John likes Mary}. 
\eeq
We encode this sentence as follows; we have:
\[
\overrightarrow{\mbox{\em John}} \in V, \quad \overrightarrow{\mbox{\em likes}} \in T, \quad \overrightarrow{\mbox{\em Mary}} \in W
\]
where we take $V$ to be the vector space spanned by men and $W$ the vector
space spanned by women. In terms of context vectors this
means that each word is its own and only context vector, which is of
course a far too simple idealisation for practical purposes.  We will
conveniently assume that all men are referred to as  {\em male}, using indices to distinguish them: {\em m$_i$}.  Thus the set of vectors
$\{\overrightarrow{\mbox{\em m}}_i\}_i$ spans $V$.  Similarly every woman will be
referred to as {\em female} and distinguished by {\em f$_j$}, for some $j$, and the set of vectors
$\{\overrightarrow{\mbox{\em f}}_j\}_j$ spans $W$.  Let us assume that {\em John}
in sentence $(\ref{sentence})$ is {\em m}$_3$ and that {\em Mary} is {\em f}$_4$. 

 If we are 
 only interested in the truth or falsity of a sentence, we have two choices in creating the sentence space $S$: it can be  spanned by two basis vectors $|0\rangle$ and $|1\rangle$ representing the truth values of {\em true} and {\em false}, or just by a
single vector $\overrightarrow{1}$, which we identify with \em
true\em,  the origin $\overrightarrow{0}$ is then identified with \em
false\em \ (so we use Dirac notation for the basis to distinguish between the origin $\overrightarrow{0}$ and the $|0\rangle$ basis vector).  This latter approach might feel a little unintuitive, but it enables us to establish a convenient connection with the relational Montague-style models of meaning, which we shall present in the last section of the paper. 

The transitive verb $\overrightarrow{\mbox{\em likes}}$ is encoded as
the \em superposition\em:
\[
\overrightarrow{\mbox{\em likes}}=\sum_{ij}\, \overrightarrow{\mbox{\em m}_i}   \otimes \overrightarrow{\mbox{\em likes}}_{ij}\otimes \overrightarrow{\mbox{\em f}_j}
\]
where $\overrightarrow{\mbox{\em likes}}_{ij}=\overrightarrow{1}$ if {\em m}$_i$ likes 
{\em f}$_j$ and $\overrightarrow{\mbox{\em likes}}_{ij}=\overrightarrow{0}$ otherwise.
Of course, in practice, the vector that we have constructed here would be obtained automatically from data using some suitable method. 

Finally, we obtain: 
\beqa
f \left(\overrightarrow{m}_3  \otimes \overrightarrow{\mbox{\em likes}} \otimes \overrightarrow{f}_4\right)
\!\!\!\!&=\!\!\!\!& 
\sum_{ij} \, \left\langle \overrightarrow{m}_3 \mid \overrightarrow{m}_i \right\rangle  \overrightarrow{\mbox{\em likes}}_{ij} 
\langle \overrightarrow{f}_j \mid \overrightarrow{f}_4 \rangle\\
\!\!\!\!&=\!\!\!\!&
\sum_{ij}\, \delta_{3i}\,\overrightarrow{\mbox{\em likes}}_{ij}\, \delta_{j4}\\
\!\!\!\!&=\!\!&
\overrightarrow{\mbox{\em likes}}_{34} \ = \ 
							\begin{cases} \overrightarrow{1} & \ \ \text{\em m}_3 \ \text{likes} \ \text{\em f}_4\\
										  \overrightarrow{0} & \ \ o.w.
							\end{cases}	
\eeqa
So we indeed obtain the correct truth-value meaning of our sentence. We are not restricted to the truth-value meaning; on the contrary, we can have, for example, degrees of meaning, as shown in section \ref{sec:comparing}.

\paragraph{Example 1b. }{\bf Two Dimensional Truth-Theoretic Meaning.} It would be more intuitive to assume that  the sentence space $S$ is spanned by two vectors $|0\rangle$ and $|1\rangle$, which stand for {\em false} and {\em true} respectively. In this case, the computing of the meaning map proceeds in exactly the same way as in the one dimensional case. The only difference is that when the sentence ``John  likes Mary" is false, the vector {\em likes}$_{ij}$ takes the value $|0\rangle$ rather than just the origin $\overrightarrow{0}$, and if it is true it takes the value $|1\rangle$ rather than $\overrightarrow{1}$.

\subsection{Negative Transitive Sentence}
The types of a sentence with negation and a transitive verb, for example ``John does not like Mary", are:
\[
n \, (n^r s j^l \sigma)\,  (\sigma^r j j^l \sigma)\, (\sigma^r j n^l)\,  n
\]
Similar to the positive case, we assume the vector spaces of the subject and object are atomic $(V,n), (W,n)$.  
The meaning space of the auxiliary verb is $(V \otimes S \otimes J \otimes V, n^r s j^l \sigma)$, that of the negation particle is $(V \otimes J \otimes J \otimes V, \sigma^r j j^l \sigma)$,  and that of the verb is $(V \otimes J \otimes W, \sigma^r j n^l)$. The `from-meaning-of-words-to-meaning-of-a-sentence' linear map $f$ is:

\noindent
\hspace{-0.5cm}\begin{minipage}{10cm}
\beqa
& f = (1_S \otimes \epsilon_J \otimes \epsilon_J) \circ (\epsilon_V \otimes 1_S \otimes 1_{J^*} \otimes \epsilon_V \otimes 1_J \otimes 1_{J^*} \otimes \epsilon_V \otimes 1_J \otimes \epsilon_W):&\\
&V \otimes (V^* \otimes S \otimes J^* \otimes V) \otimes (V^* \otimes J \otimes J^* \otimes V) \otimes (V^* \otimes J \otimes W^*) \otimes W  \to S &
\eeqa
\end{minipage}

\noindent
and depicted as:
\begin{center}
\begin{minipage}{10cm}
\hspace{-0.3cm}\epsfig{figure=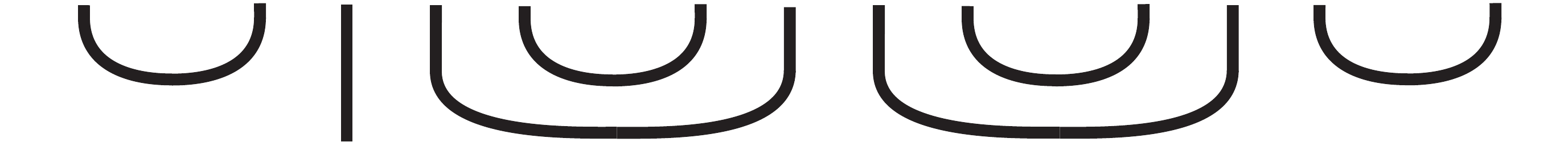,width=290pt}
\end{minipage}
\end{center}
When applied to the meaning vectors of words one obtains: 
\[
f(\overrightarrow{v} \otimes \overrightarrow{does} \otimes \overrightarrow{not} \otimes \overrightarrow{\Psi} \otimes \overrightarrow{w})
\] 
which is depicted as:
\begin{center}
\begin{minipage}{10cm}
\hspace{-0.3cm}
\epsfig{figure=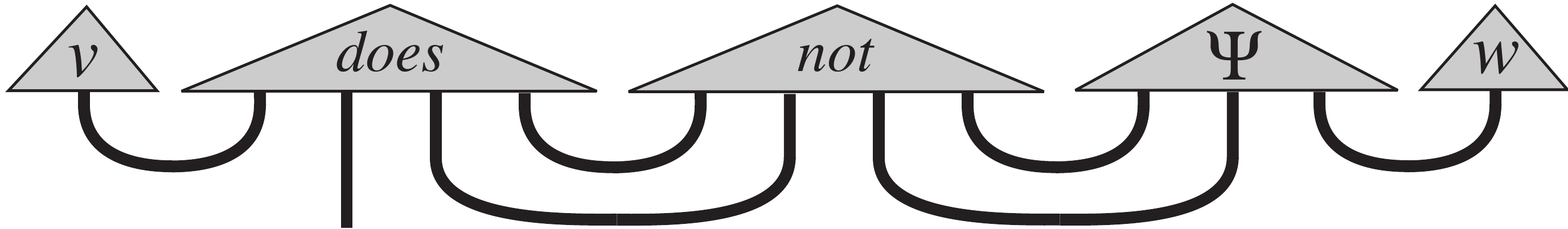,width=290pt}
\end{minipage}
\end{center} 
where $\overrightarrow{does}$ and $\overrightarrow{not}$ are the vectors corresponding to the meanings of ``does" and ``not". Since these are logical function words, we may decide to assign meaning to them manually and without consulting the document.  For instance, for {\it does} we set:
\[
S = J\qquad\mbox{and}\qquad \overrightarrow{does}=\sum_{ij}  \overrightarrow{e}_i \otimes \overrightarrow{e}_j \otimes \overrightarrow{e}_j \otimes \overrightarrow{e}_i\in V \otimes J \otimes J \otimes V \,.
\]
As explained in section \ref{sec:monoidalcats}, vectors in $V \otimes J \otimes J \otimes V $ can also be presented as linear maps of type $\mathbb{R} \to V \otimes J \otimes J \otimes V$, and in the case of {\it does} we have:
\[
\overrightarrow{does} \simeq (1_V \otimes \eta_J \otimes 1_V)\circ\eta_V: \mathbb{R} \to V \otimes J \otimes J \otimes V
\]
which shows that we only relied on structural morphisms.  

As we will demonstrate in the examples below, by relying  only on $\eta$-maps, {\it does} acts very much as an `identity' with respect to the flow of information between the words in a sentence.  This can be formalized in a more mathematical manner.
There is a well-known bijective correspondence between linear maps of type $V\to W$ and vectors in 
$V\otimes W$. Given a linear map 
$f: V\to W$ then the corresponding vector is: 
\[
\Psi_f=\sum_i \overrightarrow{e}_i\otimes f(\overrightarrow{e}_i)
\]
where $\{\overrightarrow{e}_i\}_i$ is a basis for $V$.  Diagrammatically we have:
\[
f=\raisebox{-5mm}{\epsfig{figure=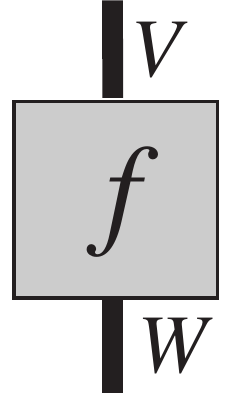,width=20pt}} \qquad\Longrightarrow\qquad \Psi_f=\raisebox{-5.1mm}{\epsfig{figure=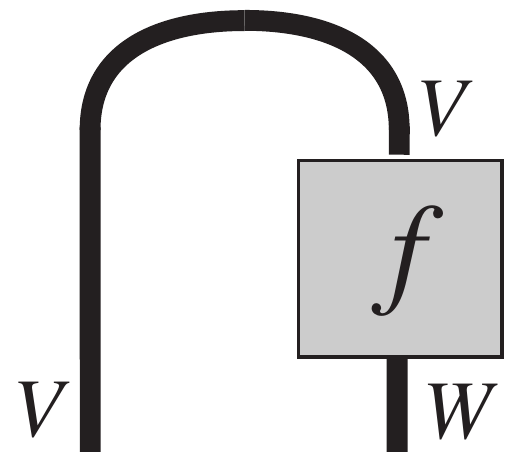,width=45pt}}
\]
Take this linear map to be the identity on $V$ and we obtain $\eta_V$.

The trick to implement {\it not} will be to take this linear map to be the linear matrix representing the logical {\it not}.  Concretely, while the matrix of the identity is $\begin{pmatrix} 1  & 0  \\ 0  & 1\end{pmatrix}$, the matrix of the logical {\it not} is $\begin{pmatrix} 0  & 1  \\ 1  & 0\end{pmatrix}$.  In Dirac notation, the vector corresponding to the identity is $|00\rangle + |11\rangle$, while the vector corresponding to the logical {\it not} is $|01\rangle + |10\rangle$.  While we have
\[
\overrightarrow{does}  \ = \  \sum_{i}  \overrightarrow{e}_i \otimes (|00\rangle + |11\rangle) \otimes \overrightarrow{e}_i
\ \in  \ V \otimes J \otimes J \otimes V\,,
\]
we will set
\[
\overrightarrow{not} \ = \ \sum_{i} \overrightarrow{e}_i \otimes (|01\rangle + |10\rangle) \otimes \overrightarrow{e}_i
\ \in  \ V \otimes J \otimes J \otimes V\,.
\]
Diagrammatically we have:
\[
\overrightarrow{does}=\raisebox{-3.2mm}{\epsfig{figure=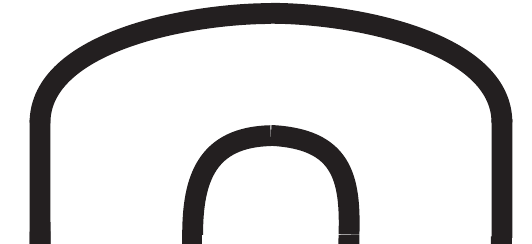,width=60pt}}\qquad\qquad \overrightarrow{not}=\raisebox{-4.5mm}{\epsfig{figure=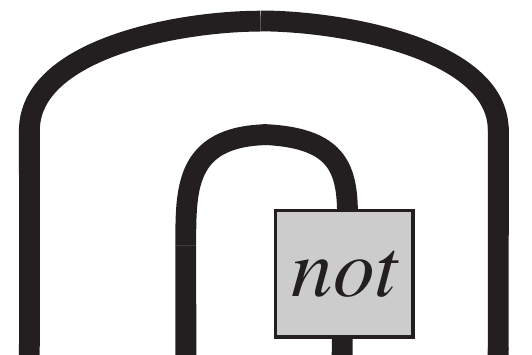,width=60pt}}
\]
Substituting all of this in $f(\overrightarrow{v} \otimes \overrightarrow{does} \otimes \overrightarrow{not} \otimes \overrightarrow{\Psi} \otimes \overrightarrow{w})$ we obtain, diagrammatically: 
\begin{center}
\hspace{-0.33cm}\epsfig{figure=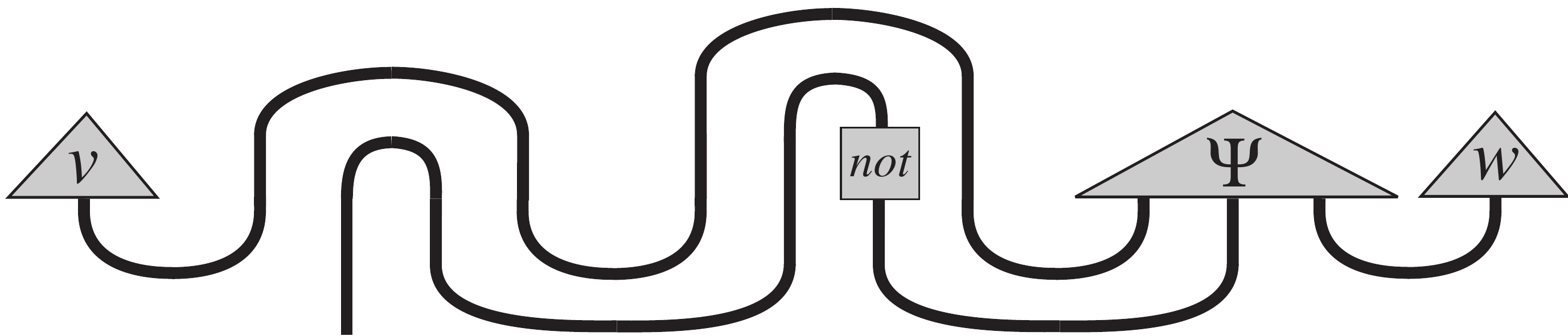,width=290pt}
\end{center} 
which by the diagrammatic calculus of compact closed categories is equal to:
\beq\label{eq:notsentence}
\hspace{-0.2cm}\epsfig{figure=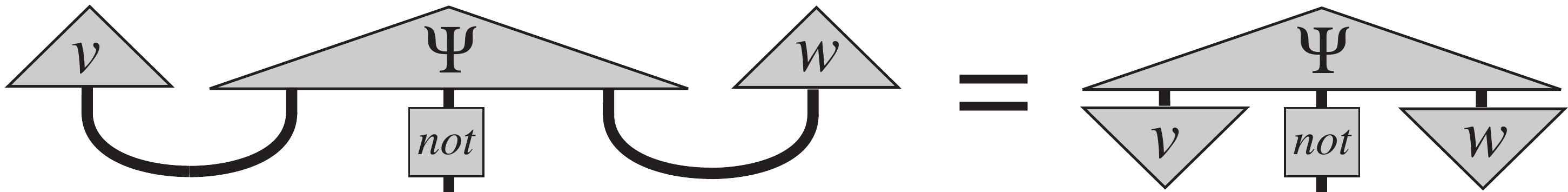,width=285pt}
\eeq 
since in particular we have that:
\begin{center}
\epsfig{figure=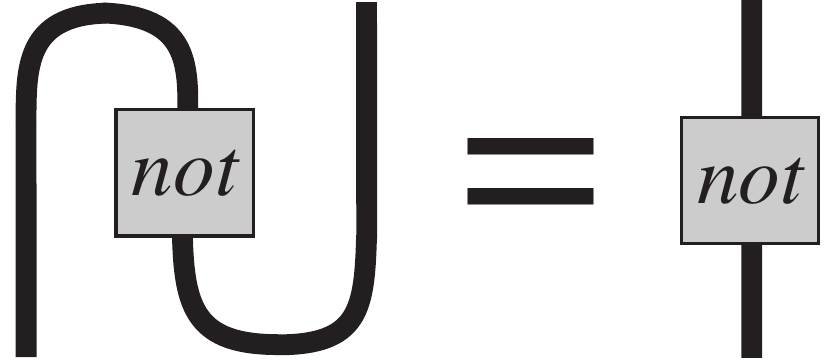,width=92pt}
\end{center} 
where the configuration on the left always encodes the transpose and the matrix of the {\it not} is obviously self-transpose. In the language of vectors and linear maps, the left hand side of eq. (\ref{eq:notsentence}) is:
\[
\left(\epsilon_V\otimes\left(\begin{array}{cc}
0 & 1 \\
1 & 0 
\end{array}\right)\otimes\epsilon_W\right)  (\overrightarrow{v} \otimes \overrightarrow {\Psi} \otimes \overrightarrow {w})\,.
\]

Note that the above pictures are very similar to the ones encountered in \cite{Kindergarten,Coe} which describe  quantum informatic protocols such as quantum teleportation and entanglement swapping.  There the morphisms $\eta$ and $\epsilon$ encode Bell-states and corresponding measurement projectors.

\paragraph{Example 2.} {\bf Negative Truth-Theoretic Meaning.}
The meaning of the sentence
\begin{center}
{\it John does not like Mary}
\end{center}
is calculated as follows.
We  assume that  the vector spaces $S = J$  are spanned by the two vectors as in Example 1b,  $|1\rangle=\begin{pmatrix} 0    \\ 1  \end{pmatrix}$ and $|0\rangle=\begin{pmatrix} 1    \\ 0 \end{pmatrix}$. We assume that $|1\rangle$ stands for \emph{true} and that $|0\rangle$ stands for  \emph{false}. 
Vector spaces $V$  and  $W$  are as in the positive case above. The vector of  {\it like}  is as before: 
\[
\overrightarrow{\mbox{\em like}}=\sum_{ij}\, \overrightarrow{\mbox{\em m}_i}  \otimes \overrightarrow{like}_{ij}\otimes \overrightarrow{\mbox{\em f}_j}
\qquad \text{for} \qquad 
\overrightarrow{like}_{ij} = \begin{cases}
|1\rangle & \mbox{\em m}_i \ \mbox{likes} \  \mbox{\em f}_j\\
|0\rangle& o.w.\end{cases}
\]
Setting $N=\left(\begin{array}{cc}
0 & 1 \\
1 & 0 
\end{array}\right)$ we obtain:
\beqa
\left(\epsilon_V\otimes N\otimes\epsilon_W\right) \left(\overrightarrow{\mbox{\em m}}_3  \otimes \overrightarrow{\mbox{\em likes}} \otimes \overrightarrow{\mbox{\em f}}_4\right)
\!\!\!\!&=\!\!\!\!&
\sum_{ij} \, \left\langle \overrightarrow{\mbox{\em m}}_3 \mid \overrightarrow{\mbox{\em m}_i} \right\rangle  N(\overrightarrow{\mbox{\em likes}}_{ij} )
\langle \overrightarrow{\mbox{\em f}}_j \mid \overrightarrow{\mbox{\em f}_4} \rangle = 
\eeqa
\[
\sum_{ij}\, \delta_{3i}\,N(\overrightarrow{\mbox{\em likes}}_{ij})\, \delta_{j4}
\quad = \quad N(\overrightarrow{\mbox{\em likes}}_{34}) =
\]
\[
\left\{\begin{array}{ll}
  |1\rangle & \ \ \overrightarrow{like}_{34}=|0\rangle\\
  |0\rangle & \ \ \overrightarrow{like}_{34}=|1\rangle
\end{array}\right.  \quad = \quad  
\left\{\begin{array}{ll} |1\rangle & \ \ \text{\em m}_3 \ \text{does not like} \ \text{\em f}_4\\
										  |0\rangle & \ \ o.w.
\end{array}\right.	
\]
That is, the meaning of ``John does not like Mary" is true if $\overrightarrow{like}_{34}$ is false, i.e. if the meaning of ``John likes Mary" is false.

For those readers who are suspicious of our graphical reasoning, here is the full-blown symbolic computation.  
Abbreviating $|10\rangle + |01\rangle$ to $\overline{n}$ and  $|00\rangle + |11\rangle$ to $\overline{d}$, and setting $f=h\circ g$ with
\[
h=1_J \otimes \epsilon_J \otimes \epsilon_J\quad\mbox{and}\quad
g=\epsilon_V \otimes 1_J \otimes 1_{J} \otimes \epsilon_V \otimes 1_J \otimes 1_{J} \otimes \epsilon_V \otimes 1_J \otimes \epsilon_W
\]
\begin{center}
\hspace{-1cm}\begin{minipage}{10cm}\small
\beqa
&&\!\!\!\!\!\!\!\!\!\!\!\!\!\!\!\!\!\!\!\!\!\!\!\!\!\!\!\!\!\!\!\!
f \left(\overrightarrow{m_3} \otimes \big(\sum_l \overrightarrow{m}_l \otimes \overline{d} \otimes  \overrightarrow{m}_l \big) \otimes
 \big(\sum_k \overrightarrow{m}_k \otimes \overline{n} \otimes \overrightarrow{m}_k \big) \otimes \big(\sum_{ij} \overrightarrow{m}_i \otimes \overrightarrow{like}_{ij} \otimes \overrightarrow{f}_j\big) \otimes \overrightarrow{f}_4 \right)\\
 &=&
h \left(\sum_{ijkl} \langle \overrightarrow{m}_3 \mid  \overrightarrow{m}_l\rangle \,\overline{d} \, \langle\overrightarrow{m}_l \mid
 \overrightarrow{m}_k \rangle\, \overline{n} \,\langle \overrightarrow{m}_k  \mid \overrightarrow{m}_i \rangle \, \overrightarrow{like}_{ij} \langle \overrightarrow{f}_j \mid \overrightarrow{f}_4  \rangle\right)\\
 &=&
h \left(\sum_{ijkl} \delta_{3l} \,\overline{d} \, \delta_{lk}\, \overline{n} \,\delta_{ki} \, \overrightarrow{like}_{ij} \delta_{j4}\right)\\
 &=&
h\left( \overline{d}   \otimes  \overline{n} \otimes  \overrightarrow{like}_{34}\right)\\
 &=&
h\left( (|00\rangle + |11\rangle) \otimes (|10\rangle + |01\rangle) \otimes   \overrightarrow{like}_{34}\right)\\
 &=&
h\left( |0 {01}{0 \overrightarrow{like}_{34}}\rangle +  |0{00}{1 \overrightarrow{like}_{34}}\rangle +
|1{11}{0 \overrightarrow{like}_{34}}\rangle + |1{10}{1 \overrightarrow{like}_{34}}\rangle\right)\\
 &=&
 |0\rangle \langle0\mid1\rangle\langle0\mid \overrightarrow{like}_{34}\rangle +  
 |0\rangle \langle0\mid0\rangle \langle1\mid \overrightarrow{like}_{34}\rangle +\\
&& |1\rangle \langle1\mid1\rangle \langle0\mid \overrightarrow{like}_{34}\rangle + 
|1\rangle \langle1\mid0\rangle \langle1\mid \overrightarrow{like}_{34}\rangle\\
 &=&
 |0\rangle  \langle1\mid \overrightarrow{like}_{34}\rangle +
|1\rangle \langle0\mid \overrightarrow{like}_{34}\rangle \\
&=&
\left\{\begin{array}{ll}
  |1\rangle & \ \ \overrightarrow{like}_{34}=|0\rangle\\
  |0\rangle & \ \ \overrightarrow{like}_{34}=|1\rangle
\end{array}\right.
\eeqa
\end{minipage}\end{center}


 \section{Comparing meanings of sentences}\label{sec:comparing}
 
One of the advantages of our approach to compositional meaning is that the meanings of sentences are all vectors in the same space, so we can use the inner product to compare the meaning vectors. This measure has been referred to and widely used as a \emph{degree of similarity} between meanings of words in the distributional approaches to meaning~\cite{Schutze}. Here we extend it to strings of words as follows.
\begin{definition}\em
Two strings of words $w_1 \cdots w_k$ \ and \ Ê$w'_1 \cdots w'_l$  \ \em have degree of similarity $m$ \em iff their Pregroup reductions result in the same grammatical type\footnote{If one wishes to do so, meaning of  phrases that do not have the same grammatical types can also be compared, but only after transferring them to a common dummy space.} and we have
\[
{1 \over N \times N'}
\left \langle f(\overrightarrow{w_1}\otimes \cdots \otimes \overrightarrow{w_k}) \mid f(\overrightarrow{w'_1}\otimes \cdots \otimes \overrightarrow{w'_l})\right \rangle = m
\]
for 
\[
N = \ \mid f(\overrightarrow{w_1}\otimes \cdots \otimes \overrightarrow{w_k})\mid \qquad
N' = \ \mid f(\overrightarrow{w'_1}\otimes \cdots \otimes \overrightarrow{w'_l}) \mid
\]
where $\mid \overrightarrow{v}\mid$ is the norm of $\overrightarrow{v}$, that is, $\mid \overrightarrow{v}\mid^2=\langle \overrightarrow{v}\mid\overrightarrow{v} \rangle$, and  $f, f'$ are the  meaning maps  defined according to definition~\ref{meaningdef}
\end{definition}
Thus we use this tool to compare meanings of positive sentences to each other, meanings of negative sentences to each other, and more importantly meanings of positive sentences to negative ones. For example, we compare the meaning of ``John  likes Mary" to ``John loves Mary", the meaning of ``John does not like Mary" to ``John does not love Mary", and also the meaning of the  latter two sentences to ``John likes Mary" and ``John loves Mary".  To make the examples more interesting, we assume that ``likes'' has degrees of both ``love'' and ``hate''.

\paragraph{Example 3.}{\bf Hierarchical Meaning.}
Similar to before,  we have:
\[
\overrightarrow{\mbox{\em loves}}=\sum_{ij}\, \overrightarrow{\mbox{\em m}_i}   \otimes \overrightarrow{\mbox{\em loves}}_{ij}\otimes \overrightarrow{\mbox{\em f}_j}
\qquad
\overrightarrow{\mbox{\em hates}}=\sum_{ij}\, \overrightarrow{\mbox{\em m}_i}   \otimes \overrightarrow{\mbox{\em hates}}_{ij}\otimes \overrightarrow{\mbox{\em f}_j}
\]
where
$\overrightarrow{\mbox{\em loves}}_{ij}= |1\rangle$ if ${\mbox{\em m}}_i$ loves 
${\mbox{\em f}}_j$ and $\overrightarrow{\mbox{\em loves}}_{ij}= |0\rangle$ otherwise, and $\overrightarrow{\mbox{\em hates}}_{ij}=|1\rangle$ if ${\mbox{\em m}_i}$ hates 
${\mbox{\em f}_j}$ and $\overrightarrow{\mbox{\em hates}}_{ij}=|0\rangle$ otherwise. 
Define {\em likes} to have degrees of {\em love} and {\em hate} as follows:
 \[
\overrightarrow{\mbox{\em likes}}={3\over 4}\overrightarrow{\mbox{\em loves}}+{1\over 4}\overrightarrow{\mbox{\em hates}} \ = \ 
\sum_{ij}\, \overrightarrow{\mbox{\em m}_i}   \otimes \left ({3\over 4} \overrightarrow{\mbox{\em loves}}_{ij} + {1 \over 4} \overrightarrow{\mbox{\em hates}}_{ij}\right)\otimes \overrightarrow{\mbox{\em f}_j}
\]
The meaning of our example sentence is thus obtained as follows:
\beqa
\!\!\!\!&&\hspace{-0.5cm}
f \left(\overrightarrow{\mbox{\em m}}_3  \otimes \overrightarrow{\mbox{\em likes}} \otimes \overrightarrow{\mbox{\em f}}_4\right) \ = \ f \left(\overrightarrow{\mbox{\em m}}_3  \otimes \left ({3\over 4}\overrightarrow{\mbox{\em loves}}+{1\over 4}\overrightarrow{\mbox{\em hates}} \right)\otimes \overrightarrow{\mbox{\em f}}_4\right)\\
\!\!\!\!&=\!\!\!\!&
\sum_{ij} \, \left\langle \overrightarrow{\mbox{\em m}}_3 \mid \overrightarrow{\mbox{\em m}_i} \right\rangle  \left ({3\over 4}\overrightarrow{\mbox{\em loves}}_{ij}+{1\over 4}\overrightarrow{\mbox{\em hates}}_{ij} \right)
\left\langle \overrightarrow{\mbox{\em f}}_j \mid \overrightarrow{\mbox{\em f}_4} \right\rangle\\
\!\!\!\!&=\!\!\!\!&
\sum_{ij}\, \delta_{3i}\,\left ({3\over 4}\overrightarrow{\mbox{\em loves}}_{ij}+{1\over 4}\overrightarrow{\mbox{\em hates}}_{ij} \right)\, \delta_{j4}\\
\!\!\!\!&=\!\!&
{3\over 4}\overrightarrow{\mbox{\em loves}}_{34}+{1\over 4}\overrightarrow{\mbox{\em hates}}_{34} 
\eeqa

 \bigskip\noindent {\bf Example 4.} {\bf Negative Hierarchical Meaning.}
To obtain the meaning of ``John does not like Mary" in this case,  one inserts ${3 \over 4} \overrightarrow{loves}_{ij} + {1 \over 4} \overrightarrow{hates}_{ij}$   for $\overrightarrow{likes}_{ij}$ in the  calculations  and 
one obtains: 
 \[
 h\left(\overline{d}   \otimes  \overline{n} \otimes  \left({3 \over 4} \overrightarrow{loves}_{34} + {1 \over 4} \overrightarrow{hates}_{34}\right) \right)
 \ = \ 
 {1\over 4}\overrightarrow{\mbox{\em loves}}_{34}+{3\over 4}\overrightarrow{\mbox{\em hates}}_{34} 
							\]
That is, the meaning of ``John does not like Mary" is the vector obtained from the meaning of ``John likes Mary" by swapping the basis vectors.  

 \bigskip\noindent {\bf Example 5.}  {\bf Degree of similarity of positive sentences.} The meanings of the distinct verbs $loves$, $likes$ and $hates$
in the different sentences propagate through the reduction mechanism 
and reveal themselves when computing inner-products between sentences in
the sentence space. For instance, the sentence ``John loves Mary" and ``John likes Mary" have a degree of similarity of 3/4, calculated as follows:
\[
\left\langle f \big (\overrightarrow{m\!}_3  \otimes \overrightarrow{\mbox{\em loves}} \otimes \overrightarrow{f}_4\big )\Bigm| f \big (\overrightarrow{m\!}_3  \otimes \overrightarrow{\mbox{\em likes}} \otimes \overrightarrow{f}_4\big )\right\rangle 
= \left\langle \overrightarrow{\mbox{\em loves}}_{34} \bigm|  \overrightarrow{\mbox{\em likes}}_{34}\right\rangle
\]
In the above, we expand the definition of $\overrightarrow{\mbox{\em likes}}_{34}$ and obtain:

\medskip
\noindent
$\left\langle \overrightarrow{\mbox{\em loves}}_{34} \bigm|  {3 \over 4} \overrightarrow{\mbox{\em loves}}_{34}
+ {1 \over 4} \overrightarrow{\mbox{\em hates}}_{34}\right\rangle =$
\[
{3 \over 4}  \left\langle \overrightarrow{\mbox{\em loves}}_{34} \bigm|  \overrightarrow{\mbox{\em loves}}_{34}
\right\rangle  +  {1 \over 4}\left\langle \overrightarrow{\mbox{\em loves}}_{34} \bigm|   \overrightarrow{\mbox{\em hates}}_{34}
\right\rangle \]
and since $\overrightarrow{\mbox{\em loves}}_{34}$ and $\overrightarrow{\mbox{\em hates}}_{34}$ are always orthogonal, that is, if one is $|1\rangle$ then the other one is $|0\rangle$, we have that
\[ 
\left\langle \overrightarrow{\mbox{\it John loves Mary}} \bigm| \overrightarrow{\mbox{\it John likes Mary}} \right\rangle
=
{3 \over 4}\mid \overrightarrow{loves}_{34} \mid^2
\] 
Hence the degree of similarity of these sentences is $\frac{3}{4}$. A similar calculation provides us with the following degrees of similarity.  For notational simplicity we drop the square of norms from now on, i.e.~we implicitly normalize meaning vectors.

\medskip
\noindent
$\left\langle \overrightarrow{\mbox{\it John hates Mary}} \bigm| \overrightarrow{\mbox{\it John likes Mary}} \right\rangle=$
\[\left\langle f \big (\overrightarrow{m\!}_3  \otimes \overrightarrow{\mbox{\em hates}} \otimes \overrightarrow{f}_4\big )\Bigm| f \big (\overrightarrow{m\!}_3  \otimes \overrightarrow{\mbox{\em likes}} \otimes \overrightarrow{f}_4\big )\right\rangle \ =  {1\over 4}\]
$\left\langle \overrightarrow{\mbox{\it John loves Mary}} \bigm| \overrightarrow{\mbox{\it John hates Mary}} \right\rangle=$
\[ \left\langle f \big (\overrightarrow{m\!}_3  \otimes \overrightarrow{\mbox{\em loves}} \otimes \overrightarrow{f}_4\big )\Bigm| f \big (\overrightarrow{m\!}_3  \otimes \overrightarrow{\mbox{\em hates}} \otimes \overrightarrow{f}_4\big )\right\rangle=0\,.\]
 
\bigskip\noindent {\bf Example 6.}  {\bf Degree of similarity of negative sentences.} In the negative case, the meaning of the composition of the meanings of the auxiliary and negation markers (``does not''),  applied to the meaning of  the verb, propagates through the computations and defines the cases of  the inner product. For instance, the sentences ``John does not love Mary" and ``John does not like Mary" have a degree of similarity of 3/4, calculated as follows:

\medskip
\noindent
$\left\langle \overrightarrow{\mbox{\it John does not love Mary}} \bigm| \overrightarrow{\mbox{\it John does not like Mary}} \right\rangle = $\\
$\left\langle f \big (\overrightarrow{m\!}_3  \otimes \overrightarrow{does} \otimes \overrightarrow{not} \otimes \overrightarrow{\mbox{\em love}} \otimes \overrightarrow{f}_4\big )\Bigm|
 f \big (\overrightarrow{m\!}_3  \otimes \overrightarrow{does} \otimes \overrightarrow{not} \otimes \overrightarrow{\mbox{\em like}} \otimes \overrightarrow{f}_4\big )\right\rangle $ \\ 
$= \left\langle \overrightarrow{\mbox{\em hates}}_{34} \Bigm|
 {1 \over 4} \overrightarrow{\mbox{\em loves}}_{34}
+ {3 \over 4} \overrightarrow{\mbox{\em hates}}_{34}\right\rangle $\\
$ = {1 \over 4} \left\langle \overrightarrow{\mbox{\em hates}}_{34} \Bigm|
 \overrightarrow{\mbox{\em loves}}_{34}\right\rangle + 
 {3 \over 4} \left\langle \overrightarrow{\mbox{\em hates}}_{34} \Bigm|
 \overrightarrow{\mbox{\em hates}}_{34}\right\rangle = {3 \over 4}$

 \bigskip\noindent {\bf Example 7.}  {\bf Degree of similarity of positive and negative sentences.} Here we compare the meanings of positive and negative sentences. This is perhaps of special interest to linguists of distributional meaning, since these sentences do not have the same grammatical structure. That we can compare these sentences shows that our approach does not limit us to the comparison of  meanings of sentences that have the same grammatical structure. We have:


\[
 \big\langle f \big (\overrightarrow{m\!}_3  \otimes \overrightarrow{\mbox{\em does}}  \otimes \overrightarrow{\mbox{\em not}} \otimes \overrightarrow{\mbox{\em like}}\otimes \overrightarrow{f}_4 \big)  \Bigm|  
  f \big(\overrightarrow{m\!}_3  \otimes \overrightarrow{\mbox{\em loves}} \otimes \overrightarrow{f}_4 \big) \big\rangle =  {1\over 4} \]
\[
 \big\langle f \big(\overrightarrow{m\!}_3  \otimes \overrightarrow{\mbox{\em does}}  \otimes \overrightarrow{\mbox{\em not}} \otimes \overrightarrow{\mbox{\em like}}\otimes \overrightarrow{f}_4 \big)   \Bigm| 
  f \big(\overrightarrow{m\!}_3  \otimes \overrightarrow{\mbox{\em hates}} \otimes \overrightarrow{f}_4 \big) \big\rangle =    {3\over 4}
\]
The following is the most interesting case:
\beqa
&&
 \big\langle f \big(\overrightarrow{m\!}_3  \otimes \overrightarrow{\mbox{\em does}}  \otimes \overrightarrow{\mbox{\em not}} \otimes \overrightarrow{\mbox{\em like}}\otimes \overrightarrow{f}_4\big)  \Bigm| 
  f \big( \overrightarrow{m\!}_3  \otimes \overrightarrow{\mbox{\em likes}} \otimes \overrightarrow{f}_4 \big) \big\rangle\\
&&=
\big\langle {1 \over 4} \overrightarrow{\mbox{\em loves}}_{34} + {3 \over 4} \overrightarrow{\mbox{\em hates}}_{34}
\Bigm|{3 \over 4} \overrightarrow{\mbox{\em loves}}_{34} + {1 \over 4} \overrightarrow{\mbox{\em hates}}_{34} \big\rangle \\
&& =
({1 \over 4} \times {3 \over 4})\big\langle  \overrightarrow{\mbox{\em loves}}_{34} \Bigm| \overrightarrow{\mbox{\em loves}}_{34}\big\rangle + 
({3 \over 4} \times {1 \over 4})\big\langle  \overrightarrow{\mbox{\em hates}}_{34} \Bigm| \overrightarrow{\mbox{\em hates}}_{34}\big\rangle\\
&& =
({1 \over 4} \times {3 \over 4}) + ({3 \over 4} \times {1 \over 4}) =
{3\over 8}
\eeqa
 This value might feel non-intuitive, since one expects that ``like" and ``does not like" have zero intersection in their meanings. This would indeed be the case had we used our original truth-value definitions. But since we have set ``like" to have degrees of   ``love" and ``hate", their intersection will no longer be 0.

Using the same method, one can form and compare meanings of many
different types of sentences.
In a full-blown vector space model, which has been automatically
extracted from large amounts of text, we obtain `imperfect' vector
representations for words, rather than the `ideal' ones presented
here.  But the mechanism of how the meanings of words propagate to the
meanings of sentences remains the same.

\section{Relations vs Vectors for Montague-style semantics} 
When fixing a base for each vector space we can think of ${\bf FVect}$ as a category of which the morphisms are matrices expressed in this base.  These matrices have real numbers as entries.  It turns out that if we consider matrices with entries not in $(\mathbb{R}, +, \times)$, but in any other semiring\footnote{A semiring is a set together with two operations, addition and multiplication, for which we have a distributive law but no additive nor multiplicative inverses.  Having an addition and multiplication of this kind suffices to have a matrix calculus.} $(R,+,\times)$,  we again obtain a compact closed category. This semiring does not have to be a field, and can for example be the positive reals $(\mathbb{R}^+, +, \times)$, positive integers $(\mathbb{N}, +, \times)$ or even Booleans $(\mathbb{B}, \vee, \wedge)$.  

In the case of $(\mathbb{B}, \vee, \wedge)$, we obtain an isomorphic copy of the category ${\bf FRel}$ of finite sets and relations with the cartesian product as tensor, as follows.  Let $X$ be a set whose elements we have enumerated as $X = \big \{x_i \mid 1 \leq i \leq\,\card{X} \big\}$.  Each element can be seen as a column  with a $1$ at the row equal to its number and $0$ in all other rows. Let $Y = \big \{y_j \mid 1 \leq j \leq\, \card{Y} \big\}$ be another enumerated set.  A relation $r \subseteq X \times Y$ is represented by  an  $\card{X}  \times \card{Y}$ matrix,
where the entry in the $i$th column and $j$th row  is $1$ iff  $(x_i, y_j) \in r$ or else $0$. 
The composite $s\circ r$ of relations $r \subseteq X \times Y$ and $s \subseteq Y \times Z$ is
\[
\{(x,z)   \mid \exists y\in Y: (x,y)\in r, (y,z)\in s  \}\,.
\]
The reader can verify that this composition induces matrix multiplication of the corresponding matrices.  

Interestingly, in  the world of relations (but not functions) there is a notion of \em superposition \em \cite{Cats}.  The relations of type $r\subseteq \{*\}\times X$ (in matricial terms, all column vectors with $0$'s and $1$'s as entries) are in bijective correspondence with the subsets of $X$ via the correspondence
\[
r\mapsto \{x\in X\mid (*,x)\in r\}\,.
\]
Each such subset can be seen as the superposition of the elements it contains.  The \em inner-product \em of two subsets is $0$ if they are disjoint and $1$ if they have a non-empty intersection.  So we can think of two disjoint sets as being \em orthogonal\em.   

Since the abstract nature of our procedure for assigning meaning to sentences did not depend on the particular choice of ${\bf FVect}$ we can now repeat it for the following situation:
\begin{diagram}
&& language &&\\
& \ldTo^{meaning} & \dTo & \rdTo^{grammar}& \\
{\bf FRel} & \lTo_{\pi_m} & {\bf FRel} \times P & \rTo_{\pi_g\!\!\!\!\!\!} & P
\end{diagram}
In ${\bf FRel}\times P$ we recover a Montague-style Boolean semantics.  The vector spaces in this setting are encodings of sets of individuals and relations over these sets. Inner products take intersections between the sets and eta maps produce new relations by connecting pairs that are not necessarily side by side.

In all our examples so far, the vector spaces of subject and object were
essentially sets that were encoded in a vector space framework. This
was done by assuming that each possible male subject is a base in the
vector space of males and similarly for the female objects. That is
why the meaning in these examples was a truth-theoretic one. We repeat
our previous calculations for example 1 in the relational setting
of ${\bf FRel}\times P$.

\bigskip\noindent {\bf Example 1 revisited.} Consider the  singleton set $\{*\}$; we assume that it  signifies the vector space $S$. We assume that the two subsets of this set, namely $\{*\}$
and $\emptyset$, will respectively identify \em true \em and
\em false\em.  We now have sets $V$, $W$ and $T=V\times \{*\}\times W$
with
\[
V:=\{\mbox{\em m}_i\}_i\,, \quad \mbox{\em likes} \subset T, \quad  W:=\{\mbox{\em f}_j\}_j
\]
such that:
\[
\mbox{\em likes}:= \{(\mbox{\em m}_i, *, \mbox{\em f}_j)\mid \mbox{\em m}_i \;\mbox{\em likes f}_j\} = \bigcup_{ij}\{\mbox{\em m}_i\}\times *_{ij}\times\{\mbox{\em f}_j\}
\]
where $*_{ij}$ is either $\{*\}$ or $\emptyset$.  So we obtain
\[
f \left(\{{\mbox{\em m}}_3\}  \times {\mbox{\em likes}} \times \{{\mbox{\em f}}_4\}\right) = 
\bigcup_{ij}  \left( \{{\mbox{\em m}}_3\}\! \cap\! \{{\mbox{\em m}}_i\} \right)  \!\times\! *_{ij} \!
\times \!
\left( \{{\mbox{\em f}}_j\}\! \cap\! \{{\mbox{\em f}_4}\} \right) = *_{34}\,.
\]

\section{Future Work}
This paper aims to lay a mathematical foundation  for the new field of  compositional distributional models of meaning in the realm of computational and mathematical  linguistics, with applications to language processing, information retrieval,  artificial intelligence, and in a conceptual way to the philosophy of language. This is just the beginning and there is so much more to do, both on the practical and the theoretical sides. Here are some examples:
\bit
\item On the logical side, our ``not"  matrix works by swapping basis and is thus essentially two dimensional. Developing a canonical matrix of negation, one that works uniformly for any dimension of the  meaning spaces,  constitutes future work. The proposal of \cite{Widdows} in using projection to the orthogonal subspace might be an option.  

\item A similar problem
arises for the meanings of other logical words, such as ``and", ``or", ``if
then". So we need to develop a general logical setting on top of our
meaning category ${\bf FVect} \times {\bf P}$. One subtlety here is that the operation that first come to mind, i.e. vector sum and product, do not correspond to logical connective of disjunction and conjunction (since e.g. they are not fully distributive). However,   the more relaxed setting of vector spaces enables us to also 
encode words such as "but", whose meaning depends on the context and thus do not have a unique logical counterpart. 

\item Formalizing the connection with Montague-semantics is another
future direction. Our above ideas can be generalized by proving a
representation theorem for ${\bf Fvect} \times {\bf P}$ on the
semiring of Booleans with respect to the category of ${\bf FRel}$ of
sets and relations. It would then be interesting to see how the so
called `non-logical' axioms of Montague are manifested at that level,
e.g.  as adjoints to substitution to recover quantifiers.

\item Along similar semantic lines, it would be good to have a
Curry-Howard-like isomorphism between non-commutative compact closed
categories, bicompact linear logic~\cite{Busz}, a version of lambda
calculus.  This will enable us to automatically obtain computations
for the meaning and type assignments of our categorical setting.

\item Our categorical axiomatics is flexible enough to accommodate
\em mixed states \em \cite{Selinger}, 
so in principle we are able to study their linguistic significance, and for instance implement the
proposals of \cite{WiddowsBruza}.

\item Finally, and perhaps most importantly, the mathematical setting
needs to be implemented and evaluated, by running experiments on real
corpus data. Efficiency and the complexity of our approach then become
an issue and need to be investigated, along with optimization
techniques. \eit

\section*{Acknowledgements}

Support from EPSRC Advanced Research Fellowship EP/D072786/1 and
European Committee grant EC-FP6-STREP 033763 for Bob Coecke, EPSRC
Postdoctoral Fellowship EP/F042728/1 for Mehrnoosh Sadrzadeh, and
EPSRC grant EP/E035698/1 for Stephen Clark are gratefully
acknowledged.  We thank Keith Van Rijsbergen,  Stephen Pulman, and Edward  Grefenstette  for
discussions, and Mirella Lapata for providing relevant references for
vector space models of meaning.

\bibliographystyle{plain}
\bibliography{LambekFestPlain}

\end{document}